 \documentclass[pmlr,twocolumn,10pt]{jmlr} 

\usepackage{booktabs}
\usepackage{siunitx}
\usepackage[bf]{caption}

\usepackage{array}
\usepackage{makecell}
\usepackage{multirow}
\usepackage{supertabular}

\newcommand{\equal}[1]{{\hypersetup{linkcolor=black}\thanks{#1}}}

\theorembodyfont{\upshape}
\theoremheaderfont{\scshape}
\theorempostheader{:}
\theoremsep{\newline}

\jmlrvolume{225}
\jmlryear{2023}
\jmlrsubmitted{LEAVE UNSET}
\jmlrpublished{LEAVE UNSET}
\jmlrworkshop{Machine Learning for Health (ML4H) 2023} 

 \title[LymphoML: Interpretable AI for Lymphoma Diagnosis]{LymphoML: An interpretable artificial intelligence-based method identifies morphologic features that correlate with lymphoma subtype}

\author{%
Vivek Shankar$^{1}$\equal{These authors contributed equally: Vivek Shankar, Xiaoli Yang, Vrishab Krishna} \quad Xiaoli Yang$^{2}$\footnotemark[1] \quad Vrishab Krishna$^{1}$\footnotemark[1] \quad Brent Tan$^3$ \quad Oscar Silva$^3$ \quad \textbf{Rebecca Rojansky}$^3$ \quad \textbf{Andrew Ng}$^1$ \quad \textbf{Fabiola Valvert}$^5$ \quad \textbf{Edward Briercheck}$^6$ \quad \textbf{David Weinstock}$^7$ \quad \textbf{Yasodha Natkunam}$^3$ \quad \textbf{Sebastian Fernandez-Pol}$^3$\equal{These authors contributed equally: Sebastian Fernandez-Pol, Pranav Rajpurkar} \textbf{Pranav Rajpurkar}$^4$\footnotemark[2] \\
$^1$\textnormal{Department of Computer Science, Stanford University}, $^2$\textnormal{Department of Statistics, Stanford University}, $^3$\textnormal{Department of Pathology, Stanford University School of Medicine}, $^4$\textnormal{Department of Biomedical Informatics, Harvard Medical School}, $^5$\textnormal{La Liga Nacional Contra el Cáncer de Guatemala (INCAN)}, $^6$\textnormal{Fred Hutchinson Cancer Research Center}, $^7$\textnormal{Dana-Farber Cancer Institute; Harvard Medical School}
\texttt{\{vivek96,xiaoliy2,vrishab,sfernand\}@stanford.edu}\\
}

\begin{document}

\maketitle

\begin{abstract}

The accurate classification of lymphoma subtypes using hematoxylin and eosin (H\&E)-stained tissue is complicated by the wide range of morphological features these cancers can exhibit. We present LymphoML - an interpretable machine learning method that identifies morphologic features that correlate with lymphoma subtypes. Our method applies steps to process H\&E-stained tissue microarray cores, segment nuclei and cells, compute features encompassing morphology, texture, and architecture, and train gradient-boosted models to make diagnostic predictions. LymphoML’s interpretable models, developed on a limited volume of H\&E-stained tissue, achieve non-inferior diagnostic accuracy to pathologists using whole-slide images and outperform black box deep-learning on a dataset of 670 cases from Guatemala spanning 8 lymphoma subtypes. Using SHapley Additive exPlanation (SHAP) analysis, we assess the impact of each feature on model prediction and find that nuclear shape features are most discriminative for DLBCL (F1-score: 78.7\%) and classical Hodgkin lymphoma (F1-score: 74.5\%). Finally, we provide the first demonstration that a model combining features from H\&E-stained tissue with features from a standardized panel of 6 immunostains results in a similar diagnostic accuracy (85.3\%) to a 46-stain panel (86.1\%). 

\end{abstract}

\begin{figure*}[!ht]
\floatconts
  {fig:Workflow}
  {\caption{\textbf{LymphoML Approach.} 
   We extract morphological, spatial, and textural features from segmented nuclei and cells in pathology images. We characterize the statistical distribution of each feature by computing summary metrics across the patch. These aggregated statistics are input features to train machine learning models to predict lymphoma subtypes.}}
  {\includegraphics[width=1.0\linewidth]{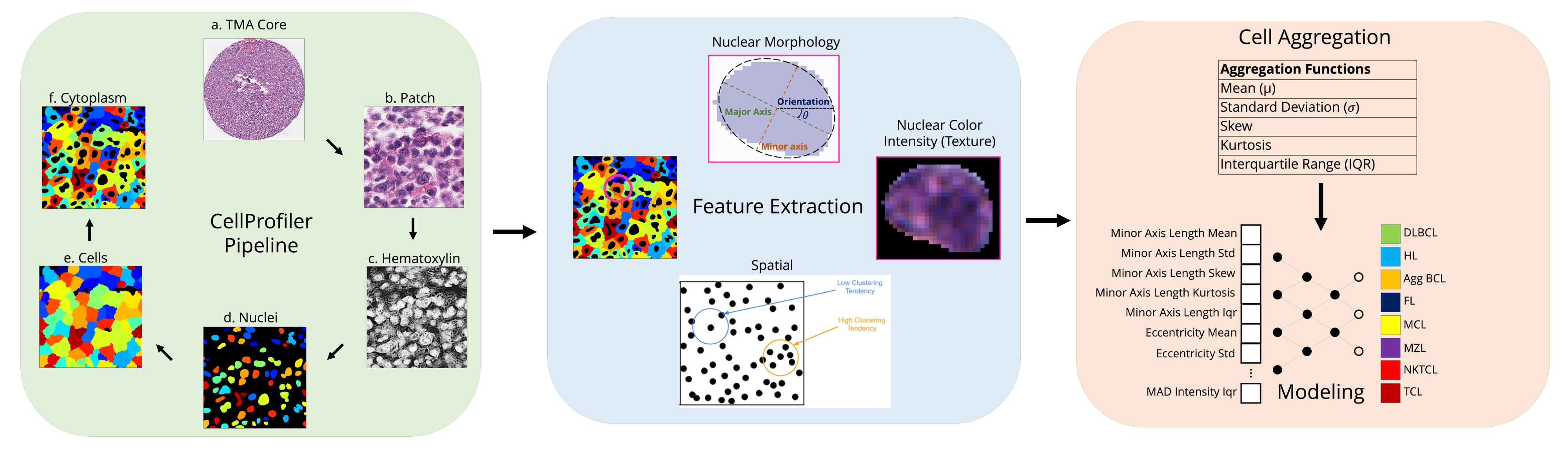}}
\end{figure*}

\section{Introduction}
\label{sec:intro}

Lymphomas are abnormal proliferations derived from lymphocytes \citep{Jamil2021-ap}. The process of precisely diagnosing lymphomas requires knowledge of the clinical history (e.g. site of involvement, history of solid organ transplant) and morphologic evaluation of hematoxylin and eosin (H\&E)-stained tissue by a trained pathologist \citep{Swerdlow2016-vu}. After evaluation of the H\&E-stained slide and relevant clinical information, one or a few diagnoses are deemed most likely to guide  ancillary testing: immunohistochemical (IHC) stains, flow cytometry, and cytogenetic and molecular studies \citep{Wang2017-kt,Sun2016-fx}. Unlike some fields of pathology, in which a definitive diagnosis is frequently possible using H\&E-stained tissue alone, for lymphoma diagnosis, IHC stains or flow cytometry are essential in most cases. This is because identifying the cell of origin for lymphomas (B-cell, T-cell, and NK cells) is essential for definitive diagnosis and treatment \citep{Nowakowski2019-ye}, but this cannot be reliably determined based on the H\&E-stained section alone. In contrast to the H\&E-stained section, which is inexpensive and widely available, IHC stains and flow cytometry require costly equipment, expensive reagents, and trained personnel. Experienced pathologists often require fewer ancillary studies, and thus greater experience may lead to more efficient resource utilization. However, worldwide, the shortage of pathologists is so great that modest improvements in the efficiency of pathologists is unlikely to make a significant impact in reducing the costs of lymphoma diagnosis \citep{Metter2019-gk, Eniu2017-kp}. Thus, strategies that can help general pathologists reduce the number of ancillary studies may help to reduce the cost of lymphoma diagnosis.

Machine learning tools applied to diagnostic pathology have shown promise in analyzing H\&E-stained images, achieving high accuracies ranging from 94-100\% when classifying between a small number of lymphomas (ranging from 2-4 diagnostic categories: diffuse large B-cell lymphoma (DLBCL), non-DLBCL \citep{Li2020-dg}; DLBCL, Burkitt lymphoma (BL) \citep{Mohlman2020-os}; DLBCL, follicular lymphoma (FL), reactive lymphoid hyperplasia \citep{Miyoshi2020-lv}; chronic lymphocytic leukemia (CLL), FL, mantle cell lymphoma (MCL) \citep{Janowczyk2016-aj, Brancati2019-nf, Zhang2020-hu}; benign, DLBCL, BL and small lymphocytic lymphoma \citep{Achi2019-ik}). However, classifying between a small number of lymphoma subtypes does not reflect the full scope of complexity encompassed in pathologist workflows. Tools that can accurately distinguish among a larger number of diagnostic categories may provide greater clinical value for diagnostic pathologists in real-world settings. Furthermore, AI tools may be useful in low-middle income countries by 1) screening specimens to reduce the number of slides that require pathologist review, 2) allowing diagnoses to be made using inexpensive and widely available H\&E-stained sections alone, or 3) allowing pathologists to maximize the diagnostic yield from the H\&E to minimize the number of IHC-stained sections necessary to make an accurate diagnosis.

Despite prior studies achieving high performance using black-box deep-learning methods, our work highlights an interpretable, feature-engineering approach for lymphoma subtyping. We hypothesized that feature-engineering methods might out-perform deep-learning due to the limited number of labeled examples for specific lymphomas in our dataset. Additionally, our model's predictions can be explained using techniques like SHapley Additive exPlanation (SHAP) analysis \citep{Lundberg2017-mi}. Model explainability is a critical component for the safety and acceptance of AI workflows in clinical practice \citep{Evans2022-gw}.

In this work, we introduce LymphoML, an interpretable machine learning approach for lymphoma subtyping into eight diagnostic categories. LymphoML segments nuclei and cells, extracts morphological, textural, and architectural features, and aggregates them into patch-level feature vectors to train classification models. LymphoML achieves an accuracy of 64.3\% on a dataset of 670 lymphoma cases from Guatemala using only H\&E-stained TMA cores and demonstrates non-inferiority to hematopathologists and general pathologists. LymphoML’s interpretable machine learning models outperform deep-learning: TripletNet (52.8\%) and ResNet (53.5\%) due to small data samples for specific diagnostic categories. Using the SHapley Additive exPlanation (SHAP) method, we find that nuclear shape features are most discriminative, especially for diffuse large B-cell lymphoma (F1-score: 78.7\%), classic Hodgkin lymphoma (F1-score: 74.5\%), and mantle cell lymphoma (F1-score: 71.0\%). Finally, combining information from the H\&E-based model with a limited set of immunohistochemical (IHC) stains results in a similar diagnostic accuracy (85.3\%) as with a larger set of IHC stains (86.1\%). Our work suggests a potential way to incorporate machine learning into clinical practice to reduce the number of expensive IHC stains while achieving similar diagnostic accuracy.

\section{Methods}
\label{sec:materials and methods}

\subsection{Dataset}
\label{sec:dataset}

The cases used for this study were selected and tissue microarrays (TMAs) were constructed as previously published \citep{Valvert2021-iz}. \citet{Valvert2021-iz} retrospectively reviewed medical records to identify formalin-fixed, paraffin-embedded (FFPE) biopsy specimens obtained at Instituto de Cancerologia y Hospital Dr. Bernardo Del Valle (INCAN) because of clinical suspicion of lymphoma between 2006 and 2018. One-half of each FFPE block was shipped to Stanford University for H\&E whole-slide image (WSI) generation. Two hematopathologists reviewed the slides, selected regions of interest (ROIs), and included two cores from each sample for tissue microarray (TMA) construction. The H\&E-stained TMAs were scanned at 40x magnification (0.25 µm per pixel) on an Aperio AT2 scanner (Leica Biosystems, Nussloch, Germany) in ScanScope Virtual Slide (SVS) format. Diagnoses were established based on the World Health Organization (WHO) classification \citep{Swerdlow2016-vu} and then binned into 8 categories: aggressive B-cell (Agg BCL), diffuse large B-cell (DLBCL), follicular (FL), classic Hodgkin (CHL), mantle cell (MCL), marginal zone (MZL), natural killer T-cell (NKTCL), or mature T-cell lymphoma (TCL). The selected categories were therapeutically driven as described in Supplemental Table 2A in \citet{Valvert2021-iz}; diagnoses that are binned together require administration of similar treatment procedures. Only a relatively small number of relevant categories were not included such as CLL, small lymphocytic leukemia, carcinoma, plasma cell neoplasm, and nonmalignant cases (reactive lymphoid hyperplasia). For a full list of the categories considered in this study and the categories excluded, see Supplemental Table 2A in \citet{Valvert2021-iz}. All of the TMA blocks (seven total) were also stained for 46-different markers by IHC stains \citep{Valvert2021-iz}. Each IHC-stained TMA was assessed by a hematopathologist to determine if the lymphoma cells were positive or negative for the marker. The complete list of cases with the associated IHC results is provided in Supplementary Table 2. The distribution of cases in each lymphoma subtype is provided in Table \ref{table1}. Of 670 FFPE biopsy specimens, 68 failed quality control (did not have sufficient tissue per core, missing ground-truth diagnoses) and were excluded from the dataset. The remaining 602 samples were split at a core-level into training, validation, and test splits with 70\% of the tissue microarray (TMA) cores for training, 10\% for validation to tune hyperparameters, and 20\% for testing. Stratified sampling was used to proportionally represent the eight diagnostic categories in each of the training, validation, and test sets (\figureref{fig:DataSplits}).

\subsection{Patch Extraction}
\label{sec:patch extraction}

The H\&E-stained tissue cores were indicated by hematopathologists using Qupath \citep{Bankhead2017-nd}. From each tissue core, we extracted a fixed number of non-overlapping patches at 40x magnification, starting from the top-left and proceeding until the bottom-right corner. We omitted patches that were mostly white and contained little tissue. Specifically, background was defined as pixels with saturation value less than 0.05 in HSV space, and we excluded patches where more than 95\% of the pixels were background.

\subsection{Nuclei and Cell Segmentation}
\label{sec:nuclei and cell segmentation}

We considered two different deep-learning based nuclear segmentation models: HoVer-Net \citep{Graham2019-af} and StarDist \citep{Schmidt_2018} to segment every nucleus inside the H\&E-stained TMA cores. HoVer-Net uses a neural network based on a pre-trained ResNet-50 architecture to extract image features. StarDist is powered by a pre-trained deep-learning CNN that predicts a suitable shape representation (star-convex polygon) for each cell nucleus. We normalized the input image pixel intensities to the range 0.0 to 1.0 using percentiles of 1 and 99 to clip the bottom and top 1\% of pixel values to 0.0 and 1.0. Then, we ran StarDist, which operated independently on each TMA core and produced an output image segmenting all individual cell nuclei in the core. 
We selected StarDist as the nuclei segmentation algorithm for all our cases. We measured the agreement of HoVer-Net’s and StarDist's nuclei segmentations by computing the mean Intersection over Union (mIOU) over all segmented patches. We obtained a mIOU of 0.762. Additionally, we found that the best-performing H\&E-only models utilizing features extracted from StarDist achieved marginally higher top-1 accuracy (64.3\%) than the best-performing models using features extracted from HoVer-Net (61.5\%).

\subsection{Feature Extraction}
\label{sec:feature extraction}

We used the per-nucleus binary segmentation masks output by StarDist to compute geometric features for each cell nucleus using methods similar to those by \cite{Vrabac2021-zg}. Using manually extracted features allowed our models to produce interpretable results and facilitated identification of features that were most important in driving the classification using SHAP (SHapley Additive exPlanations, described below) \citep{Lundberg2017-mi}. We calculated features such as Feret diameters, convex hull area of the segmented nucleus, and derived geometric features including measures of circularity, elongation, and convexity. To obtain a richer feature set, we used CellProfiler \citep{Carpenter2006-en}, an open-source tool for analyzing biological images, to extract quantitative features of the morphology, color intensity, and texture of segmented nuclei and cells. We constructed an image analysis pipeline in CellProfiler consisting of modules to process the H\&E cores, identify nuclear and cell boundaries, and measure features of the identified objects (\figureref{fig:Workflow}). First, a color deconvolution was performed on each patch to create separate hematoxylin and eosin-stained images in grayscale. Next, we ran StarDist on the hematoxylin image to produce a binary mask segmenting the nuclei. The nuclei were subsequently used as a reference to identify secondary objects such as the cells and cytoplasm. Finally, we extracted size, shape (e.g. bounding box area, minor axis length), color intensity (e.g. mean intensity, integrated intensity), and textural features from the detected cells and nuclei. The full list of features extracted by CellProfiler is provided in Table \ref{table3.2} \citep{Stirling2021-nr}. To obtain a single feature vector for each patient, each of the features was aggregated across all nuclei in a patch by their mean, standard deviation, skew, kurtosis, and percentiles, yielding a total of 1595 features for each patient.

\paragraph{Spatial Relationship Features.}
\label{sec:spatial relationship features}

To model spatial relationships between nuclei, we considered architectural features from two sources: 1) CPArch: features that contain architectural information provided by CellProfiler (BoundingBoxMaximum, Center), and 2) CT: spatial features representing clustering tendency (CT) using Ripley’s K function. We followed the steps described in \citet{Subramanian2018-no} to compute CT. We used centroid coordinates (in pixels) to define cell locations in each patch and computed values of the self-K function at different radii. The optimal radii range was determined by cross-validation. The resulting vector consisting of self-K function values at each radius was used as the patch’s CT feature. When performing lymphoma subtype predictions, we concatenated CPArch and CT vectors directly with the rest of the features.

\subsection{Models}
\label{sec:models}

We used LightGBM \citep{Ke2017-kg}, a tree-based machine learning algorithm that employs a gradient boosting framework. We handled class-imbalance by preserving the label distribution when splitting the dataset and made sure patches from the same patient were in the same data split. To correct the bias induced by class-imbalance during model training, we used focal loss \citep{Lin2017-xw} and turned on ‘balanced’ mode in LightGBM to adjust weights inversely proportional to class frequencies in the input data. We used 5-fold cross-validation for all experiments. We experimented with hyperparameter tuning on the number of leaves, maximum depth, and number of epochs for gradient-boosting models.

For deep-learning models, we divided cores into patches of 224x224 pixels with 50\% overlap data augmentation and filtered patches as described in “Patch Extraction.” Patch pixels were normalized to have mean 0, variance 1. We fine-tuned two open-source models pretrained on H\&E patches – a ResNet-50 self-supervised on several tasks and cancers with H\&E and IHC-stained slides \citep{He2016-uo} and a specialized TripletNet architecture pre-trained on CAMELYON16 (dataset of breast cancer H\&E WSIs) \citep{Srinidhi2022-yi}. We experimented with hyperparameter tuning on the learning rate (in the range: 1e-2, 1e-3, 1e-4, 1e-5) and unfreezing different numbers of layers of the pre-trained TripletNet and ResNet while fine-tuning. The deep-learning results we reported use the best identified hyperparameters (learning rate: 0.001, allowing weights in all layers to update during fine-tuning) on the validation set. Focal loss was used to handle class-imbalance with normalized weights generated from label proportions on the training set and a gamma parameter of 2.0. The model was updated by an Adam Optimizer \citep{Kingma2014-ie} with a learning rate of 0.001 and batch size of 128 for 100 epochs.

\subsection{Feature Importances}
\label{sec:feature importances}

We used the SHapley Additive exPlanation (SHAP) method to quantify the impact of each feature on the trained model \citep{Lundberg2017-mi}. The SHAP method explains prediction by allocating credit among input features; feature credit is calculated using Shapley Values as the change in the expected value of the model's predicted score for a label when a feature is present versus absent. We also grouped related morphological features into different categories and ran SHAP on the feature groups. We summed the raw SHAP values within each group to estimate the group’s importance.

\subsection{Evaluation}
\label{sec:evaluation}

We assessed the performance of our models (index test) in predicting the ground-truth WHO diagnosis (reference standard) for each case. The assessors of the reference standard reviewed H\&E slides and IHC results to classify each specimen according to the WHO classification \citep{Valvert2021-iz}. We additionally compared our models to human-benchmark pathologists. The H\&E-stained TMA cores/WSIs were reviewed by hematopathologists who were blinded to all other clinical data including immunohistochemical stains and the final histopathologic diagnosis (reference standard result). Other than the H\&E-stained tissue, no additional clinical information were provided to the human-benchmark pathologist or to our models.

We analyzed model performance by top-1 accuracy, F1 score, AUROC, sensitivity, and specificity. Top-1 accuracy, the proportion of examples for which the predicted label matches the target label, provides a direct measure of model/pathologist performance and was used in several prior works including \citet{Li2020-dg} and \citet{Steinbuss}. We calculated metrics for each label individually and their weighted average across all labels. We weight by the support, the number of true instances for each label, to account for class imbalance. Let $k$ represent the total number of labels, $n$ represent the total number of examples, and $|y_i|$  represent the number of examples with label $i$. We compute the weighted F1 score:

\begin{equation}
\text{Weighted\_F1\_Score} = \frac{1}{n} \sum_{i=1}^{k} |y_{i}| * \text{F1\_Score}_{i}
\end{equation}

We followed a similar procedure to calculate weighted sensitivity, specificity, and AUROC metrics. We computed 95\% CIs using non-parametric bootstrapping from 1,000 bootstrap samples. These metrics are summarized in Table \ref{table6.1} for models using features extracted from H\&E-stained tissue only. We performed paired t-tests checking for any significant differences and equivalence (TOST) relationships by comparing the top-1 accuracy of the Best H\&E Model with pathologists on H\&E-stained TMA cores and WSIs. The two-tailed paired t-test between the Best H\&E Model and Pathologist checks whether the mean difference for a particular metric (e.g. top-1 accuracy) between the two methods is zero. The significance level is 5\%. Let $\mu_{\text{Best H\&E Model}}$ represent the mean of a metric (e.g. mean top-1 accuracy) for the Best H\&E Model and $\mu_{\text{Pathologist}}$ represent the mean of a metric for the Pathologist. TOST consists of two one-tailed paired t-tests with the following null hypothesis:

\begin{equation}
|\mu_{\text{Best H\&E Model}} - \mu_{\text{Pathologist}}| \geq +\Delta 
\end{equation}

Statistical test results are summarized in Table \ref{table9}. This study was conducted in compliance with STARD guidelines (Supplementary Table 1). 

\section{Results}
\label{sec:results}

\subsection{Nuclear Morphology}
\label{sec:nuclear morphology}

We first tested whether nuclear shape features had higher diagnostic yield than nuclear texture or cytoplasmic features for classifying lymphoma subtypes. The model using only nuclear features achieved 59.7\% ([51.2\%, 68.2\%]) top-1 test accuracy, while models using nuclear texture or cytoplasmic features alone achieved slightly lower accuracy (Table \ref{table6.1}). Adding nuclear texture or cytoplasmic features to the nuclear shape features marginally improved performance by 1-2\%. We analyzed model performance by lymphoma subtype using per-class F1 scores. Nuclear features were most discriminative for diffuse large B-cell lymphoma (DLBCL) (F1 score: 76.2\%), classic Hodgkin lymphoma (CHL) (F1 score: 65.3\%), and mantle cell lymphoma (MCL) (F1 score: 51.6\%) (Table \ref{table6.3}). To extract morphologic features from H\&E-stained images, we used the data processing, feature extraction, and modeling workflow described in \figureref{fig:Workflow-Original}. 

\paragraph{Feature Importances.}
\label{sec:feature importances 2}

We utilized the SHAP method to investigate whether area shape features played the biggest role in classifying lymphoma subtypes among all nuclear features. We reported the resulting Shapley values for the top 20 nuclear features on individual predictions. We also trained a parsimonious model using only the top eight most impactful features for each class as determined by SHAP; the resulting model achieved a top-1 test accuracy of 61.2\% ([53.4\%, 69.0\%]) using just 10\% of all features. The majority of the top 20 nuclear features were area shape features such as mean radius, minor axis length, maximum feret diameter, solidity, orientation, maximum radius length, and nuclei area.

For select features, we analyzed their ability to distinguish patients with specific lymphoma subtypes. For example, the MinorAxisLength parameter was significantly different between cases of DLBCL and MCL (\figureref{fig:MALComparison}). The minor axis length (MAL) is the length (in pixels) of the ellipse that has the same normalized second central moments as the nucleus region \citep{Stirling2021-nr}. DLBCL generally has cells with a larger minor axis length, consistent with the WHO definition of DLBCL as a B-cell lymphoma with large cells \citep{Swerdlow2016-vu}. We grouped related morphological features into different categories and ran SHAP on the resulting feature groups. The nuclear size feature group had the largest mean absolute SHAP value (\figureref{fig:SHAPGrouped}), suggesting that of all nuclear features, size features were most helpful for classifying DLBCL, CHL, and MCL.

\subsection{Nuclear Architecture}
\label{sec:nuclear architecture}

We hypothesized that incorporating architectural features such as clustering tendency among nuclei would improve accuracy on certain types of lymphoma over nuclear morphological features alone. The best model utilizing architectural and nuclear features was “Nuclear Morphological + Cytoplasm + Intensity + CPArch'' (referred to as “Best H\&E Model” below). It achieved 64.3\% ([55.7\%, 72.9\%]) top-1 accuracy, the highest accuracy among all models using H\&E features, though statistical tests did not suggest it being significantly different from the Nuclear Morphological Model. Although Best H\&E Model, which uses all inputs except CT, slightly outperforms the model using all available features, the difference between their performances is not statistically significant (\figureref{fig:PerfSummary}). We examined Best H\&E Model's per-class performance to find that it achieved 71.0\% ([55.0\%, 87.0\%]) F1 score in predicting MCL, 19.4\% higher than the Nuclear Morphological Model’s F1 score for MCL (51.6\% [32.2\%, 71.0\%]), though this difference did not achieve statistical significance.

\subsection{Comparison to Black-Box Features}
\label{sec:comparison to black-box features}

We hypothesized that the interpretable machine learning models produced by LymphoML would outperform deep-learning methods (TripletNet and ResNet) used in prior studies given the scarcity of labeled examples. There was no significant difference between the test performance of the two deep-learning methods. ResNet, which achieved better test accuracy than TripletNet, was significantly inferior to the Best H\&E Model. It was also worse than the Nuclear Morphological Model in both test accuracy and F1 score by a margin of
{\raise.17ex\hbox{$\scriptstyle\sim$}}5\% (top-1 test accuracy of TripletNet: 52.8\% [44.2\%, 61.4\%], top-1 test accuracy of ResNet: 53.5\% [44.8\%, 62.2\%]). For each class, the baseline generally had better performances.

\subsection{Pathologist Comparison}
\label{sec:pathologist comparison}

We compared the performance of the Best H\&E Model from Section \ref{sec:nuclear architecture} with pathologists on H\&E-stained TMA cores and WSIs (\figureref{fig:PerfComparison}). Best H\&E Model’s test accuracy (64.3\% [55.7\%, 72.9\%]) surpassed all pathologists (Table \ref{model-pathologist-overall}). Statistical tests verified that Best H\&E Model was non-inferior to the General Pathologist on WSIs and Hematopathologist 1 on TMAs, and there was no evidence of its performance being statistically different from any pathologist’s performance (Table \ref{table9}). A similar conclusion can be derived from other metrics (sensitivity, specificity, and AUROC). We compared the per-class performance of our methods. On one hand, compared to pathologists’ predictions, the Best H\&E Model failed to effectively identify any case in MZL and TCL while pathologists generally achieved 30\% and 23.5\% F1 scores in diagnosing MZL and TCL respectively. On the other hand, Best H\&E Model achieved a 71.0\% F1 score ([55.0\%, 87.0\%]) in MCL, surpassing all hematopathologists by a margin \textgreater18\% (Table \ref{model-pathologist-per-class}) and statistically superior to Hematopathologist 1 on H\&E TMAs and Hematopathologist 3 on WSIs. Best H\&E Model achieves consistent and better performance than hematopathologists only for the 3 categories of DLBCL, CHL, and MCL for which we have a sufficiently large number of cases in our cohort.

\begin{table*}[]
\begin{tabular}{c|c|c|c|c|c}
\hline
\multirow{2}{*}{\textbf{Method}} &
  \multirow{2}{*}{\textbf{\begin{tabular}[c]{@{}c@{}}Accuracy\end{tabular}}} &
  \multirow{2}{*}{\textbf{\begin{tabular}[c]{@{}c@{}}Sensitivity\end{tabular}}} &
  \multirow{2}{*}{\textbf{\begin{tabular}[c]{@{}c@{}}Specificity\end{tabular}}} &
  \multirow{2}{*}{\textbf{\begin{tabular}[c]{@{}c@{}}AUROC\end{tabular}}} &
  \multirow{2}{*}{\textbf{\begin{tabular}[c]{@{}c@{}}F1 Score\end{tabular}}} \\
 &
   &
   &
   &
   &
   \\ \hline
\begin{tabular}[c]{@{}c@{}}Hematopathologist 1 on H\&E TMAs\end{tabular} &
  \begin{tabular}[c]{@{}c@{}}$56.1 \pm 8.1$\end{tabular} &
  \begin{tabular}[c]{@{}c@{}}$56.5 \pm 9.7$\end{tabular} &
  \begin{tabular}[c]{@{}c@{}}$82.1 \pm 5.4$\end{tabular} &
  N/A &
  \begin{tabular}[c]{@{}c@{}}$53.5 \pm 8.5$\end{tabular} \\ 
\begin{tabular}[c]{@{}c@{}}Hematopathologist 2 on H\&E TMAs\end{tabular} &
  \begin{tabular}[c]{@{}c@{}}$60.1 \pm 7.5$\end{tabular} &
  \begin{tabular}[c]{@{}c@{}}$60.5 \pm 8.2$\end{tabular} &
  \begin{tabular}[c]{@{}c@{}}$82.8 \pm 5.4$\end{tabular} &
  N/A &
  \begin{tabular}[c]{@{}c@{}}$58.8 \pm 8.4$\end{tabular} \\ 
\begin{tabular}[c]{@{}c@{}}Hematopathologist 3 on WSIs\end{tabular} &
  \begin{tabular}[c]{@{}c@{}}$63.5 \pm 7.4$\end{tabular} &
  \begin{tabular}[c]{@{}c@{}}$63.9 \pm 9.9$\end{tabular} &
  \begin{tabular}[c]{@{}c@{}}$92.9 \pm 2.8$\end{tabular} &
  N/A &
  \begin{tabular}[c]{@{}c@{}}$66.0 \pm 8.2$\end{tabular} \\ 
\begin{tabular}[c]{@{}c@{}}General Pathologist on WSIs\end{tabular} &
  \begin{tabular}[c]{@{}c@{}}$56.1 \pm 7.4$\end{tabular} &
  \begin{tabular}[c]{@{}c@{}}$55.8 \pm 9.7$\end{tabular} &
  \begin{tabular}[c]{@{}c@{}}$93.2 \pm 1.9$\end{tabular} &
  N/A &
  \begin{tabular}[c]{@{}c@{}}$65.1 \pm 7.3$\end{tabular} \\
\begin{tabular}[c]{@{}c@{}}Best H\&E Model\end{tabular} &
  \begin{tabular}[c]{@{}c@{}}$64.3 \pm 8.6$\end{tabular} &
  \begin{tabular}[c]{@{}c@{}}$66.9 \pm 6.0$\end{tabular} &
  \begin{tabular}[c]{@{}c@{}}$88.7 \pm 2.6$\end{tabular} &
  \begin{tabular}[c]{@{}c@{}}$85.9 \pm 2.9$\end{tabular} &
  \begin{tabular}[c]{@{}c@{}}$58.5 \pm 9.5$\end{tabular} \\ \hline
\end{tabular}
\caption{Overall class-weighted performance metrics of Best H\&E Model vs pathologists using TMAs/WSIs.}
\label{model-pathologist-overall}
\end{table*}

\begin{table*}
\centering
\begin{tabular}{c|c|c|c|c}
\hline
\textbf{Method}                  & \textbf{DLBCL} & \textbf{HL} & \textbf{Agg BCL} & \textbf{FL} \\ \hline
Hematopathologist 1 on H\&E TMAs & $73.3 \pm 7.8$     & $63.8 \pm 14.9$ & $25.0 \pm 25.0$      & $35.7 \pm 23.6$  \\ 
Hematopathologist 2 on H\&E TMAs & $73.3 \pm 7.4$ & $86.4 \pm 9.4$  & 0.0 & $42.9 \pm 21.6$        \\ 
Hematopathologist 3 on WSIs      & $82.1 \pm 6.6$ & $86.4 \pm 9.4$  & 0.0 & $58.3 \pm 24.5$ \\ 
General Pathologist on WSIs      & $67.8 \pm 9.3$ & $80.0 \pm 12.3$ & 0.0 & $54.5 \pm 21.4$ \\ 
Best H\&E Model                  & $78.7 \pm 7.7$ & $74.5 \pm 13.4$ & 0.0 & $31.6 \pm 24.8$         \\ \hline
\textbf{Method} & \textbf{MCL} & \textbf{MZL} & \textbf{NKTCL} & \textbf{TCL} \\ \hline
Hematopathologist 1 on H\&E TMAs & $36.4 \pm 23.6$  & $28.6 \pm 28.6$  & 0.0            & $23.5 \pm 23.5$  \\ 
Hematopathologist 2 on H\&E TMAs & $43.5 \pm 23.2$ & $17.4 \pm 17.4$ & $26.7 \pm 26.7$ & 0.0         \\ 
Hematopathologist 3 on WSIs & $20.0 \pm 20.0$ & $30.8 \pm 30.7$ & $70.0 \pm 20.0$ & $23.5 \pm 23.5$ \\ 
General Pathologist on WSIs & $52.2 \pm 23.7$ & $50.0 \pm 38.9$ & 1.0 & $23.5 \pm 23.5$ \\
Best H\&E Model & $71.0 \pm 16.0$ & 0.0 & $25.0 \pm 25.0$ & 0.0 \\ \hline
\end{tabular}
\caption{Per-class F1 score comparison of Best H\&E Model to pathologists using TMAs/WSIs.}
\label{model-pathologist-per-class}
\end{table*}

\begin{figure}[htb]
\floatconts
  {fig:PerfSummary}
  {\caption{Performance summary for all H\&E models with Majority Selection line representing a baseline that always predicts the most prevalent class and Random Selection line representing a baseline that performs random classification. The best performing model (“Best H\&E Model”) is marked with a star.}}
  {\includegraphics[width=0.8\linewidth]{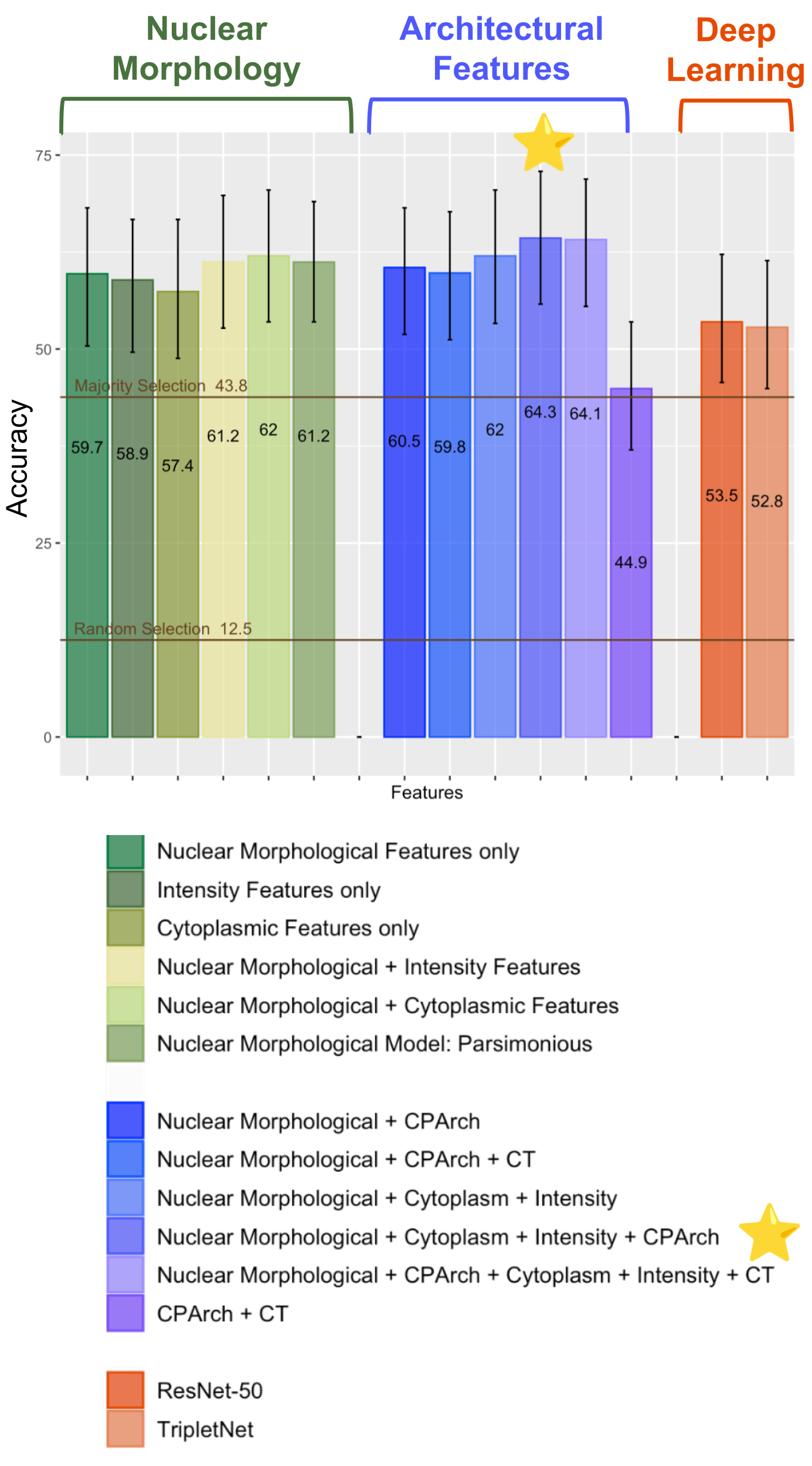}}
\end{figure}

\subsection{Additional Stains}
\label{sec:additional stains}

The ground-truth diagnosis for our cohort was based on review by hematopathologists of the H\&E-stained whole-slide images and a panel of 46 immunohistochemical (IHC) stains that were performed on all cases. For each case and immunostain, a hematopathologist scored the lymphoma cells as positive or negative for the immunostain (e.g. CD30-positive) or “cannot interpret.” We set out to determine the diagnostic accuracy of using this IHC information alone, using a limited set of IHC stain results, and using a limited set of IHC stain results in conjunction with the H\&E-based model.

We considered six candidate immunostains based on our impression of markers that would provide the highest yield considering the diagnostic categories in our cohort: CD10, CD20, CD3, EBV-ISH, BCL1/cyclin D1, and CD30. For each immunostain, the pathologist’s score was included as an additional categorical feature to the Best H\&E Model. We also grouped lymphoma subtypes into five categories that are similar in terms of clinical behavior and therapeutic approaches: B-cell lymphomas (DLBCL, Agg BCL), CHL, FL and MZL, MCL, and T-cell lymphomas (NKTCL, TCL). We evaluated the accuracy and F1 scores of models using features extracted from H\&E stains along with different combinations of immunostain indicator features. The baseline model using 46 immunostains (and no H\&E) achieved a top-1 accuracy of 86.1\% ([80.0\%, 92.2\%]). Using the six selected immunostains alone, without the H\&E, the model achieved an accuracy of 75.2\% ([68.2\%, 82.2\%]), statistically inferior to the model using all 46 immunostains (\figureref{fig:PerfComparison}). The Best H\&E Model augmented by six immunostains (CD10, CD20, CD3, EBV-ISH, BCL1, CD30) achieved an accuracy of 85.3\% ([79.9\%, 90.7\%]) and showed no evidence of difference from the model using all 46 immunostains.

\begin{figure}[!ht]
\floatconts
  {fig:PerfComparison}
  {\caption{Performance comparison between hematopathologists and the best-performing H\&E model (left). Performance comparisons of models using features extracted from H\&E and selected immunostains (right).}}
  {\includegraphics[width=1.0\linewidth]{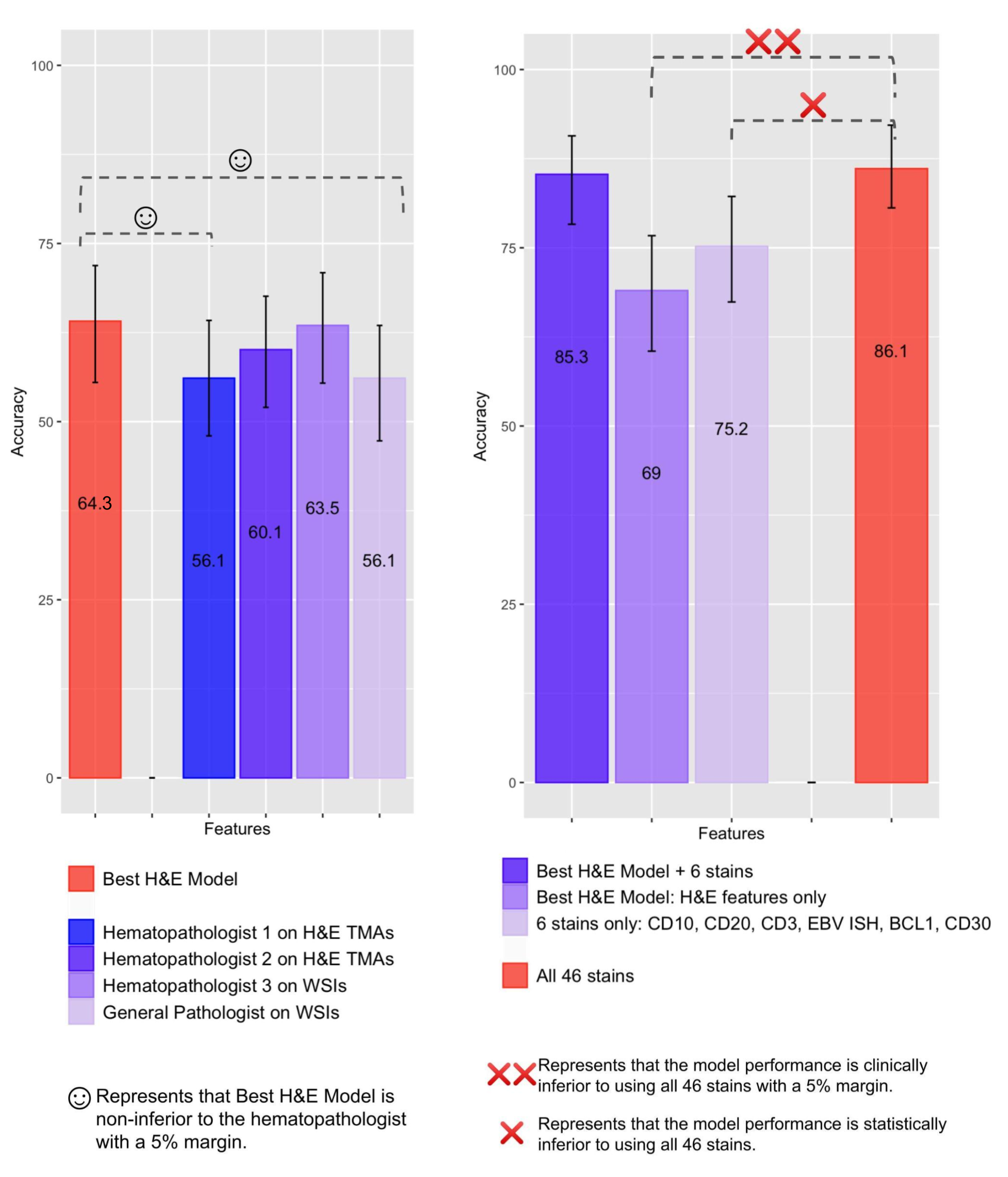}}
\end{figure}

\section{Discussion}
\label{sec:discussion}

In this study, we assessed the performance of interpretable and deep-learning approaches in the classification of eight lymphoma categories using a cohort of 670 lymphoma cases. Using only H\&E-stained TMA material, LymphoML achieves performance comparable to that of experienced hematopathologists reviewing only the H\&E-stained tissue. The highest performance was achieved using an interpretable model. Combining information from the H\&E-based model with a limited set of IHC stains resulted in a similar diagnostic accuracy as with a much larger set of IHC stains. It is notable that our interpretable models, developed on a limited volume of tissue, achieved the same diagnostic accuracy as hematopathologists using whole-slide images. Our study suggests that computational tools can extract more diagnostic information from less tissue than a pathologist. The growing use of needle core biopsies in clinical practice makes judicious use of tissue even more critical. Computational tools that maximize the diagnostic yield of the H\&E-stained slide could potentially reduce the number of ancillary stains and thus avoid the need for repeat biopsy or an excisional biopsy. 

\paragraph{Feature engineering effective on limited data.} Prior studies that applied machine learning methods to lymphoma diagnosis have achieved accuracies of 94\% to 100\%, but it is not clear how these tools can be employed in the real-world given the limited number of diagnostic categories studied. Our study was performed on a cohort that contains 2- to 4-fold more diagnostic categories than prior studies. Though our cohort contains the second-largest number of cases thus far used to develop machine learning tools for lymphoma diagnosis, the number of cases is within the same order of magnitude as that used in other studies. In real-world settings with a large number of possible diagnoses, deep-learning methods will require a proportionate increase in the number of cases to perform well. With a limited number of examples for specific diagnostic categories, models can only learn a small range of the wide morphological variability present in these subtypes. Therefore, feature-engineering approaches may yet provide superior diagnostic yield when the number of cases per diagnosis is insufficient to use deep-learning. Furthermore, we note that deep-learning models will have comparable performance to prior work (e.g. DLBCL vs non-DLBCL by \citet{Li2020-dg}) if they are limited to the same number of classes and provided a sufficiently large number of samples per class. 

\paragraph{Nuclear features align with pathologist review.} Pathologists use features similar to the nuclear shape features identified by SHAP including nuclear-to-cytoplasmic ratio, nuclear contour irregularities, and nuclear size as one of many clues to determine if cells are normal or malignant. DLBCL, by definition, consists of sheets of large B-cells, with ‘large’ defined as nuclei that are at least the size of a histiocyte nucleus or two-times the size of a normal lymphocyte. Consistent with this definition used to render the ground-truth diagnosis, our interpretable model showed that nuclear size features differed between DLBCL and other lymphoma subtypes. Specific features that provided the greatest diagnostic yield included mean radius, minor axis length, maximum Feret diameter, solidity, and orientation.

\paragraph{Combining assessment with immunostains can be cost-effective.} In our study, we provide the first demonstration that a model that incorporates immunophenotypic features from six selected IHC stains and features extracted from H\&E-stained sections achieves the same diagnostic accuracy as a model that uses a larger number of immunostains. As H\&E-stained slides are cheaper than IHC stains by at least an order of magnitude, extracting the maximal diagnostic yield from H\&E can reduce the number of IHC stains ordered and costs without a reduction in diagnostic accuracy. Additionally, we note that pathologists usually order a custom panel of IHC stains for each case. In our study, we showed that models combining H\&E features with a standard set of six IHC stains can arrive at the correct diagnosis (85.3\%). Standardizing IHCs reduces variability in practice and improves cost effectiveness.

\paragraph{Strengths.}
\label{sec:strengths}

The best LymphoML models achieved top-1 test accuracy, sensitivity, and specificity equivalent to that of experienced hematopathologists. Using feature importance analysis, our computational methods help pathologists better understand the primary characteristics of the lesion that contribute to the model’s prediction. Our study cohort from Guatemala, a population that is not represented in current digital pathology datasets, will prove valuable in the effort to build diverse datasets for computational tools in medicine. Most prior works focused on the use of whole-slide images, often with expensive, manually-generated patch-level annotations. WSIs often require manual annotations because these larger images usually contain both cancerous and normal surrounding tissue. Here, we use TMAs, which do not require expensive manual annotations because cores are already enriched for lymphoma. Using TMAs, computational tools are more cost-effective and usable in low-middle income countries where acquisition of labeled data can prove costly.

\paragraph{Limitations.}
\label{sec:limitations}

First, the data were collected and processed in a single institution by a single slide scanner. We should evaluate the model’s generalizability on cohorts from other institutions collected using different technical setups and slides scanned on different machines. Next, TMAs capture only a small portion of the full tumor volume and are much smaller than WSIs \citep{Beck2011-bu}. Thus, we could have likely trained more powerful models by analyzing WSIs. The heavy class imbalance meant that only a small number of examples were available for Agg BCL, MZL, TCL (\textless10 patients), which can only display a small subset of the wide morphological variability present in these subtypes. Finally, to implement LymphoML in a clinical setting, pathologists would have to suspect one of the validated diagnostic categories, select a region of interest, and then the model would render a favored diagnosis. Future studies with more diagnostic categories, and a more diverse set of background tissues may someday enable automated identification of target lesion(s) and subsequent diagnostic categorization.

\section*{Data and Code Availability}

The datasets used and/or analyzed during the current study are available from Sebastian Fernandez-Pol (sfernand@stanford.edu) on reasonable request. We are acquiring a Data Use Agreement (DUA) to be able to share the raw data. Our code is made available on Github at \href{https://github.com/rajpurkarlab/LymphoML}{this link}. 

\section*{Author Contributions}

V.S., X.Y., V.K., S.F., and P.R. developed the concept and design; V.S., X.Y., V.K., S.F., P.R., B.T., O.S., R.R., F.V., E.B., D.W., and Y.N. performed acquisition, analysis, or interpretation of data; A.N., S.F., and P.R. provided supervision. V.S., X.Y., V.K., S.F., and P.R. drafted the manuscript, and all authors provided critical revision of manuscript for important intellectual content.

\section*{Ethics}

This study was approved by the institutional review boards of the Dana-Farber Cancer Institute and Stanford University and the ethics committee of La Liga Nacional Contra el Cáncer. 

\bibliography{shankar23}

\begin{thebibliography}{35}
\providecommand{\natexlab}[1]{#1}
\providecommand{\url}[1]{\texttt{#1}}
\expandafter\ifx\csname urlstyle\endcsname\relax
  \providecommand{\doi}[1]{doi: #1}\else
  \providecommand{\doi}{doi: \begingroup \urlstyle{rm}\Url}\fi

\bibitem[Achi et~al.(2019)Achi, Belousova, Chen, Wahed, Wang, Hu, Kanaan, Rios, and Nguyen]{Achi2019-ik}
Hanadi~El Achi, Tatiana Belousova, Lei Chen, Amer Wahed, Iris Wang, Zhihong Hu, Zeyad Kanaan, Adan Rios, and Andy N~D Nguyen.
\newblock Automated diagnosis of lymphoma with digital pathology images using deep learning.
\newblock \emph{Ann. Clin. Lab. Sci.}, 49\penalty0 (2):\penalty0 153--160, March 2019.

\bibitem[Bankhead et~al.(2017)Bankhead, Loughrey, Fern{\'a}ndez, Dombrowski, McArt, Dunne, McQuaid, Gray, Murray, Coleman, James, Salto-Tellez, and Hamilton]{Bankhead2017-nd}
Peter Bankhead, Maurice~B Loughrey, Jos{\'e}~A Fern{\'a}ndez, Yvonne Dombrowski, Darragh~G McArt, Philip~D Dunne, Stephen McQuaid, Ronan~T Gray, Liam~J Murray, Helen~G Coleman, Jacqueline~A James, Manuel Salto-Tellez, and Peter~W Hamilton.
\newblock {QuPath}: Open source software for digital pathology image analysis.
\newblock \emph{Sci. Rep.}, 7\penalty0 (1):\penalty0 16878, December 2017.

\bibitem[Beck et~al.(2011)Beck, Sangoi, Leung, Marinelli, Nielsen, van~de Vijver, West, van~de Rijn, and Koller]{Beck2011-bu}
Andrew~H Beck, Ankur~R Sangoi, Samuel Leung, Robert~J Marinelli, Torsten~O Nielsen, Marc~J van~de Vijver, Robert~B West, Matt van~de Rijn, and Daphne Koller.
\newblock Systematic analysis of breast cancer morphology uncovers stromal features associated with survival.
\newblock \emph{Sci. Transl. Med.}, 3\penalty0 (108):\penalty0 108ra113, November 2011.

\bibitem[Brancati et~al.(2019)Brancati, De~Pietro, Frucci, and Riccio]{Brancati2019-nf}
Nadia Brancati, Giuseppe De~Pietro, Maria Frucci, and Daniel Riccio.
\newblock A deep learning approach for breast invasive ductal carcinoma detection and lymphoma {Multi-Classification} in histological images.
\newblock \emph{IEEE Access}, 7:\penalty0 44709--44720, 2019.

\bibitem[Carpenter et~al.(2006)Carpenter, Jones, Lamprecht, Clarke, Kang, Friman, Guertin, Chang, Lindquist, Moffat, Golland, and Sabatini]{Carpenter2006-en}
Anne~E Carpenter, Thouis~R Jones, Michael~R Lamprecht, Colin Clarke, In~Han Kang, Ola Friman, David~A Guertin, Joo~Han Chang, Robert~A Lindquist, Jason Moffat, Polina Golland, and David~M Sabatini.
\newblock {CellProfiler}: image analysis software for identifying and quantifying cell phenotypes.
\newblock \emph{Genome Biol.}, 7\penalty0 (10):\penalty0 R100, October 2006.

\bibitem[Eniu et~al.(2017)Eniu, Martei, Trimble, and Shulman]{Eniu2017-kp}
Alexandru~E Eniu, Yehoda~M Martei, Edward~L Trimble, and Lawrence~N Shulman.
\newblock Cancer care and control as a human right: Recognizing global oncology as an academic field.
\newblock \emph{Am Soc Clin Oncol Educ Book}, 37:\penalty0 409--415, 2017.

\bibitem[Evans et~al.(2022)Evans, Retzlaff, Gei{\ss}ler, Kargl, Plass, M{\"u}ller, Kiehl, Zerbe, and Holzinger]{Evans2022-gw}
Theodore Evans, Carl~Orge Retzlaff, Christian Gei{\ss}ler, Michaela Kargl, Markus Plass, Heimo M{\"u}ller, Tim-Rasmus Kiehl, Norman Zerbe, and Andreas Holzinger.
\newblock The explainability paradox: Challenges for {xAI} in digital pathology.
\newblock \emph{Future Gener. Comput. Syst.}, 133:\penalty0 281--296, August 2022.

\bibitem[Graham et~al.(2019)Graham, Vu, Raza, Azam, Tsang, Kwak, and Rajpoot]{Graham2019-af}
Simon Graham, Quoc~Dang Vu, Shan E~Ahmed Raza, Ayesha Azam, Yee~Wah Tsang, Jin~Tae Kwak, and Nasir Rajpoot.
\newblock {Hover-Net}: Simultaneous segmentation and classification of nuclei in multi-tissue histology images.
\newblock \emph{Med. Image Anal.}, 58:\penalty0 101563, December 2019.

\bibitem[Gupta et~al.(2010)Gupta, Gupta, Singh, Gupta, and Kudesia]{Gupta2010-ff}
Shilpa Gupta, Ruchika Gupta, Sompal Singh, Kusum Gupta, and Madhur Kudesia.
\newblock Nuclear morphometry and texture analysis of b-cell non-hodgkin lymphoma: utility in subclassification on cytosmears.
\newblock \emph{Diagn. Cytopathol.}, 38\penalty0 (2):\penalty0 94--103, February 2010.

\bibitem[He et~al.(2016)He, Zhang, Ren, and Sun]{He2016-uo}
Kaiming He, Xiangyu Zhang, Shaoqing Ren, and Jian Sun.
\newblock Deep residual learning for image recognition.
\newblock In \emph{Proceedings of the {IEEE} conference on computer vision and pattern recognition}, pages 770--778, 2016.

\bibitem[Jamil and Mukkamalla(2021)]{Jamil2021-ap}
Ayesha Jamil and Shiva Kumar~R Mukkamalla.
\newblock Lymphoma.
\newblock In \emph{{StatPearls}}. StatPearls Publishing, Treasure Island (FL), September 2021.

\bibitem[Janowczyk and Madabhushi(2016)]{Janowczyk2016-aj}
Andrew Janowczyk and Anant Madabhushi.
\newblock Deep learning for digital pathology image analysis: A comprehensive tutorial with selected use cases.
\newblock \emph{J. Pathol. Inform.}, 7:\penalty0 29, July 2016.

\bibitem[Ke et~al.(2017)Ke, Meng, Finley, Wang, Chen, Ma, Ye, and Liu]{Ke2017-kg}
Guolin Ke, Qi~Meng, Thomas Finley, Taifeng Wang, Wei Chen, Weidong Ma, Qiwei Ye, and Tie-Yan Liu.
\newblock {LightGBM}: A highly efficient gradient boosting decision tree.
\newblock \emph{Adv. Neural Inf. Process. Syst.}, 30, 2017.

\bibitem[Kingma and Ba(2014)]{Kingma2014-ie}
Diederik~P Kingma and Jimmy Ba.
\newblock Adam: A method for stochastic optimization.
\newblock December 2014.

\bibitem[Lesty et~al.(1986)Lesty, Raphael, Nonnenmacher, Leblond-Missenard, Delcourt, Homond, and Binet]{Lesty1986-ze}
C~Lesty, M~Raphael, L~Nonnenmacher, V~Leblond-Missenard, A~Delcourt, A~Homond, and J~L Binet.
\newblock An application of mathematical morphology to analysis of the size and shape of nuclei in tissue sections of non-hodgkin's lymphoma.
\newblock \emph{Cytometry}, 7\penalty0 (2):\penalty0 117--131, March 1986.

\bibitem[Li et~al.(2020)Li, Bledsoe, Zeng, Liu, Hu, Bi, Liang, and Li]{Li2020-dg}
Dongguang Li, Jacob~R Bledsoe, Yu~Zeng, Wei Liu, Yiguo Hu, Ke~Bi, Aibin Liang, and Shaoguang Li.
\newblock A deep learning diagnostic platform for diffuse large b-cell lymphoma with high accuracy across multiple hospitals.
\newblock \emph{Nat. Commun.}, 11\penalty0 (1):\penalty0 6004, November 2020.

\bibitem[Lin et~al.(2017)Lin, Goyal, Girshick, He, and Doll{\'a}r]{Lin2017-xw}
Tsung-Yi Lin, Priya Goyal, Ross Girshick, Kaiming He, and Piotr Doll{\'a}r.
\newblock Focal loss for dense object detection.
\newblock In \emph{Proceedings of the {IEEE} international conference on computer vision}, pages 2980--2988, 2017.

\bibitem[Lundberg and Lee(2017)]{Lundberg2017-mi}
Scott~M Lundberg and Su-In Lee.
\newblock A unified approach to interpreting model predictions.
\newblock \emph{Adv. Neural Inf. Process. Syst.}, 30, 2017.

\bibitem[Metter et~al.(2019)Metter, Colgan, Leung, Timmons, and Park]{Metter2019-gk}
David~M Metter, Terence~J Colgan, Stanley~T Leung, Charles~F Timmons, and Jason~Y Park.
\newblock Trends in the {US} and canadian pathologist workforces from 2007 to 2017.
\newblock \emph{JAMA Netw Open}, 2\penalty0 (5):\penalty0 e194337, May 2019.

\bibitem[Miyoshi et~al.(2020)Miyoshi, Sato, Kabeya, Yonezawa, Nakano, Takeuchi, Ozawa, Higo, Yanagida, Yamada, Kohno, Furuta, Muta, Takeuchi, Sasaki, Yoshimura, Matsuda, Muto, Moritsubo, Inoue, Suzuki, Sekinaga, and Ohshima]{Miyoshi2020-lv}
Hiroaki Miyoshi, Kensaku Sato, Yoshinori Kabeya, Sho Yonezawa, Hiroki Nakano, Yusuke Takeuchi, Issei Ozawa, Shoichi Higo, Eriko Yanagida, Kyohei Yamada, Kei Kohno, Takuya Furuta, Hiroko Muta, Mai Takeuchi, Yuya Sasaki, Takuro Yoshimura, Kotaro Matsuda, Reiji Muto, Mayuko Moritsubo, Kanako Inoue, Takaharu Suzuki, Hiroaki Sekinaga, and Koichi Ohshima.
\newblock Deep learning shows the capability of high-level computer-aided diagnosis in malignant lymphoma.
\newblock \emph{Lab. Invest.}, 100\penalty0 (10):\penalty0 1300--1310, October 2020.

\bibitem[Mohlman et~al.(2020)Mohlman, Leventhal, Hansen, Kohan, Pascucci, and Salama]{Mohlman2020-os}
Jeffrey~S Mohlman, Samuel~D Leventhal, Taft Hansen, Jessica Kohan, Valerio Pascucci, and Mohamed~E Salama.
\newblock Improving augmented human intelligence to distinguish burkitt lymphoma from diffuse large {B-Cell} lymphoma cases.
\newblock \emph{Am. J. Clin. Pathol.}, 153\penalty0 (6):\penalty0 743--759, May 2020.

\bibitem[Nowakowski et~al.(2019)Nowakowski, Feldman, Rimsza, Westin, Witzig, and Zinzani]{Nowakowski2019-ye}
Grzegorz~S Nowakowski, Tatyana Feldman, Lisa~M Rimsza, Jason~R Westin, Thomas~E Witzig, and Pier~Luigi Zinzani.
\newblock Integrating precision medicine through evaluation of cell of origin in treatment planning for diffuse large b-cell lymphoma.
\newblock \emph{Blood Cancer J.}, 9\penalty0 (6):\penalty0 48, May 2019.

\bibitem[Schmidt et~al.(2018)Schmidt, Weigert, Broaddus, and Myers]{Schmidt_2018}
Uwe Schmidt, Martin Weigert, Coleman Broaddus, and Gene Myers.
\newblock Cell detection with star-convex polygons.
\newblock In \emph{Medical Image Computing and Computer Assisted Intervention {\textendash} {MICCAI} 2018}, pages 265--273. Springer International Publishing, 2018.
\newblock \doi{10.1007/978-3-030-00934-2_30}.
\newblock URL \url{https://doi.org/10.1007%2F978-3-030-00934-2_30}.

\bibitem[Srinidhi et~al.(2022)Srinidhi, Kim, Chen, and Martel]{Srinidhi2022-yi}
Chetan~L Srinidhi, Seung~Wook Kim, Fu-Der Chen, and Anne~L Martel.
\newblock Self-supervised driven consistency training for annotation efficient histopathology image analysis.
\newblock \emph{Med. Image Anal.}, 75:\penalty0 102256, January 2022.

\bibitem[Steinbuss et~al.(2021)Steinbuss, Kriegsmann, Zgorzelski, Brobeil, Goeppert, Dietrich, Mechtersheimer, and Kriegsmann]{Steinbuss}
Georg Steinbuss, Mark Kriegsmann, Christiane Zgorzelski, Alexander Brobeil, Benjamin Goeppert, Sascha Dietrich, Gunhild Mechtersheimer, and Katharina Kriegsmann.
\newblock Deep learning for the classification of non-hodgkin lymphoma on histopathological images.
\newblock \emph{Cancers}, 13:\penalty0 2419, 05 2021.
\newblock \doi{10.3390/cancers13102419}.

\bibitem[Stirling et~al.(2021)Stirling, Swain-Bowden, Lucas, Carpenter, Cimini, and Goodman]{Stirling2021-nr}
David~R Stirling, Madison~J Swain-Bowden, Alice~M Lucas, Anne~E Carpenter, Beth~A Cimini, and Allen Goodman.
\newblock {CellProfiler} 4: improvements in speed, utility and usability.
\newblock \emph{BMC Bioinformatics}, 22\penalty0 (1):\penalty0 433, September 2021.

\bibitem[Subramanian et~al.(2018)Subramanian, Tang, Chidester, Ma, and Do]{Subramanian2018-no}
Vaishnavi Subramanian, Weizhao Tang, Benjamin Chidester, Jian Ma, and Minh~N Do.
\newblock Integration of spatial distribution in {Imaging-Genetics}.
\newblock In \emph{Medical Image Computing and Computer Assisted Intervention -- {MICCAI} 2018}, pages 245--253. Springer International Publishing, 2018.

\bibitem[Sun et~al.(2016)Sun, Medeiros, and Young]{Sun2016-fx}
Ruifang Sun, L~Jeffrey Medeiros, and Ken~H Young.
\newblock Diagnostic and predictive biomarkers for lymphoma diagnosis and treatment in the era of precision medicine.
\newblock \emph{Mod. Pathol.}, 29\penalty0 (10):\penalty0 1118--1142, October 2016.

\bibitem[Swerdlow et~al.(2016)Swerdlow, Campo, Pileri, Harris, Stein, Siebert, Advani, Ghielmini, Salles, Zelenetz, and Jaffe]{Swerdlow2016-vu}
Steven~H Swerdlow, Elias Campo, Stefano~A Pileri, Nancy~Lee Harris, Harald Stein, Reiner Siebert, Ranjana Advani, Michele Ghielmini, Gilles~A Salles, Andrew~D Zelenetz, and Elaine~S Jaffe.
\newblock The 2016 revision of the world health organization classification of lymphoid neoplasms.
\newblock \emph{Blood}, 127\penalty0 (20):\penalty0 2375--2390, May 2016.

\bibitem[Valvert et~al.(2021)Valvert, Silva, Sol{\'o}rzano-Ortiz, Puligandla, Sili{\'e}zar~Tala, Guyon, Dixon, L{\'o}pez, L{\'o}pez, Car{\'\i}as~Alvarado, Terbrueggen, Stevenson, Natkunam, Weinstock, and Briercheck]{Valvert2021-iz}
Fabiola Valvert, Oscar Silva, Elizabeth Sol{\'o}rzano-Ortiz, Maneka Puligandla, Marcos~Mauricio Sili{\'e}zar~Tala, Timothy Guyon, Samuel~L Dixon, Nelly L{\'o}pez, Francisco L{\'o}pez, C{\'e}sar~Camilo Car{\'\i}as~Alvarado, Robert Terbrueggen, Kristen~E Stevenson, Yasodha Natkunam, David~M Weinstock, and Edward~L Briercheck.
\newblock Low-cost transcriptional diagnostic to accurately categorize lymphomas in low- and middle-income countries.
\newblock \emph{Blood Adv}, 5\penalty0 (10):\penalty0 2447--2455, May 2021.

\bibitem[Vrabac et~al.(2021)Vrabac, Smit, Rojansky, Natkunam, Advani, Ng, Fernandez-Pol, and Rajpurkar]{Vrabac2021-zg}
Damir Vrabac, Akshay Smit, Rebecca Rojansky, Yasodha Natkunam, Ranjana~H Advani, Andrew~Y Ng, Sebastian Fernandez-Pol, and Pranav Rajpurkar.
\newblock {DLBCL-Morph}: Morphological features computed using deep learning for an annotated digital {DLBCL} image set.
\newblock \emph{Sci Data}, 8\penalty0 (1):\penalty0 135, May 2021.

\bibitem[Wang and Zu(2017)]{Wang2017-kt}
Huan-You Wang and Youli Zu.
\newblock Diagnostic algorithm of common mature {B-Cell} lymphomas by immunohistochemistry.
\newblock \emph{Arch. Pathol. Lab. Med.}, 141\penalty0 (9):\penalty0 1236--1246, September 2017.

\bibitem[Yu et~al.(2021)Yu, Li, Wang, Yeh, and Chuang]{Yu2021}
Wei-Hsiang Yu, Chih-Hao Li, Renching Wang, Chao-Yuan Yeh, and Shih-Sung Chuang.
\newblock Machine learning based on morphological features enables classification of primary intestinal t-cell lymphomas.
\newblock \emph{Cancers}, 13:\penalty0 5463, 10 2021.
\newblock \doi{10.3390/cancers13215463}.

\bibitem[Zhang et~al.(2020)Zhang, Cui, Guo, Wang, and Wang]{Zhang2020-hu}
Jianfei Zhang, Wensheng Cui, Xiaoyan Guo, Bo~Wang, and Zhen Wang.
\newblock Classification of digital pathological images of non-hodgkin's lymphoma subtypes based on the fusion of transfer learning and principal component analysis.
\newblock \emph{Med. Phys.}, 47\penalty0 (9):\penalty0 4241--4253, September 2020.

\bibitem[Zhu et~al.(2022)Zhu, Hazoglou, Li, and Dogan]{Zhu2022}
Menglei Zhu, Michael Hazoglou, Anyi Li, and Ahmet Dogan.
\newblock Automatic triaging of hematopathology tissue specimens by neural network on whole slide image (wsi).
\newblock \emph{Laboratory Investigation}, 102:\penalty0 1058--59, March 2022.

\end{thebibliography}

\clearpage

\appendix

\counterwithin{figure}{section}
\begin{figure*}[htbp]
\section{Supplementary Figures}\label{apd:first}
\floatconts
  {fig:Workflow-Original}
  {\caption{\textbf{LymphoML Approach.} 
   Patches of a fixed size are extracted from each TMA core. We used the StarDist algorithm to produce a nuclei segmentation mask. Using the H image and the nuclei segmentation mask together, we used CellProfiler to identify the cell boundaries. We identified the cytoplasm by ``subtracting" the nuclei objects from the cell objects. For each identified nucleus, cell, and cytoplasm, the following groups of features were extracted and measured: (a) nuclear morphology features, (b) spatial/architectural features (c) texture/intensity features. For each measurement obtained, the mean, standard deviation, skew, kurtosis, and interquartile range (IQR) were calculated for the entire population of objects present in a patch. The aggregated features were packed into a patch-level feature vector. Using the feature vector as input, models were trained to predict the most likely lymphoma subtype for the patch. The plurality vote across the patch-level model predictions was used to obtain the final core-level prediction.}}
  {\includegraphics[width=0.95\linewidth]{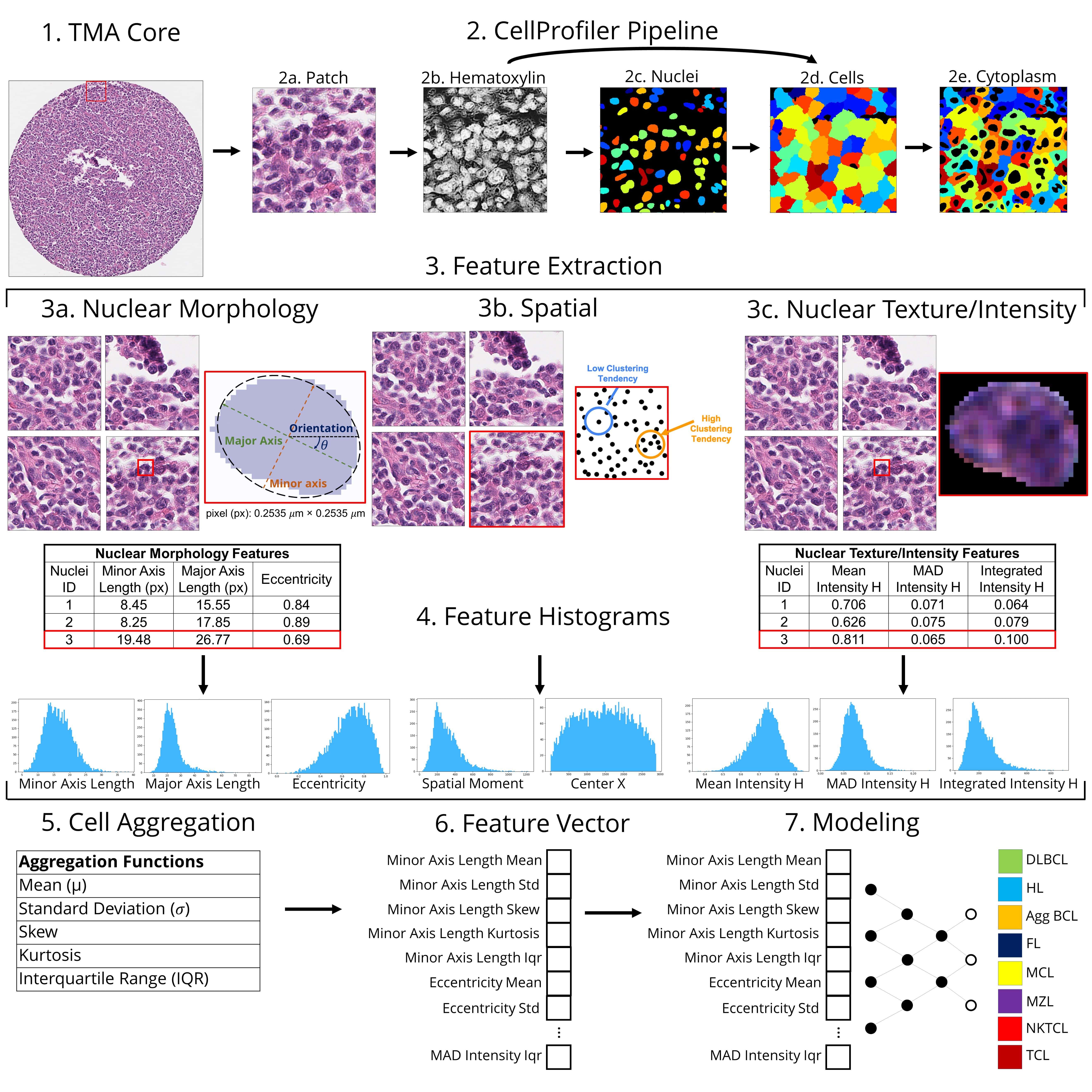}}
\end{figure*}

\begin{figure*}[htbp]
\floatconts
  {fig:MALComparison}
  {\caption{Comparison of Minor Axis Length (MAL) across lymphoma subtypes. For selected cases from (i) DLBCL, (ii) CHL/HL, and (iii) MCL, we plot a heatmap superimposed on the tissue microarray (TMA) cores showing the variability in nuclei minor axis length across the core. We also mark the patches with the largest (1) and smallest (2) mean MAL. For each patch, we display the segmented nuclei using StarDist. In (iv), we show the distribution of nuclei MAL in the maximum (1) and minimum (2) patches for each case. In (v), we show the distribution of nuclei MAL in the overall cores. In (vi), we show the distribution of mean MAL across cores in the entire TMA. DLBCL cases generally have the largest nuclei and greater variability in nuclei size than either CHL or MCL cases. In (vii), we display a waterfall plot of the mean minor axis length across patients. For each case/patient, we plot the mean of the top five patches with the highest minor axis length. DLBCL cases generally have the largest mean minor axis lengths of all lymphoma subtypes. In (viii), we show a Kernel density estimation plot. The distribution of minor axis length is clearly further to the right for DLBCL than for CHL and MCL. In (ix), we show a feature importance analysis plot by SHAP values. The top 10 nuclear features by percentage importance using SHAP are shown: minor axis length interquartile range (IQR) is the second-most important feature. All of the top 10 features are area-shape features.}}
  {\includegraphics[width=0.95\linewidth]{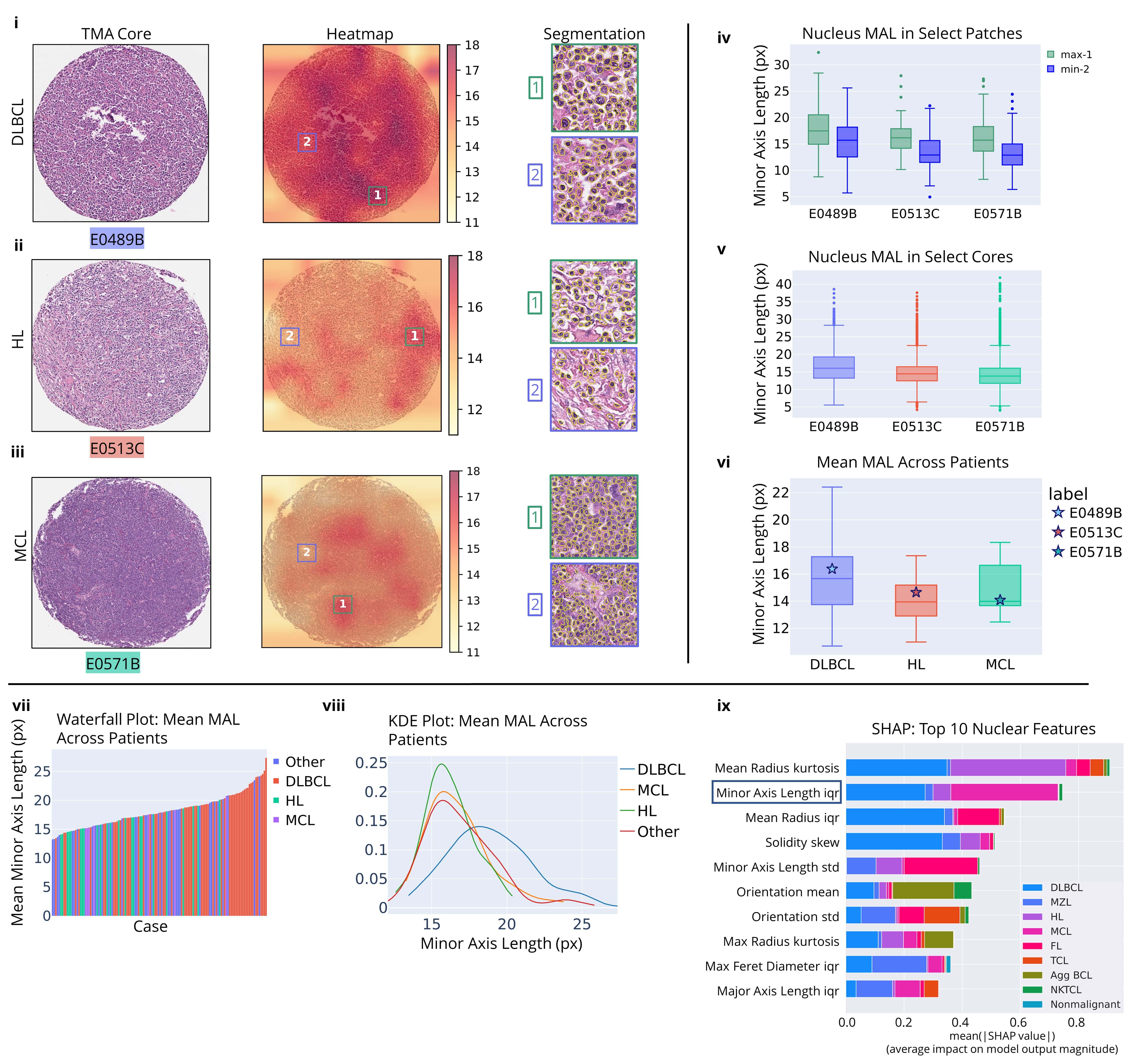}}
\end{figure*}

\begin{figure*}[htbp]
\floatconts
  {fig:DataSplits}
  {\caption{Of 670 FFPE biopsy specimens, 68 failed quality control (e.g. did not have sufficient tissue per core, at least two full samples per patient, were missing ground-truth diagnoses, missing immunohistochemical stains, etc.) and were excluded from the dataset used to train and evaluate the model. The remaining 602 samples were split at a core-level into training, validation, and test splits to ensure that all extracted image patches from the same patient are in the same data split with 70\% of the total tissue microarray (TMA) cores for training, 10\% for validation to tune model hyperparameters, and 20\% for testing. Stratified sampling was used to proportionally represent the eight diagnostic categories in each of the training, validation, and test sets.}}
  {\includegraphics[width=1.0\linewidth]{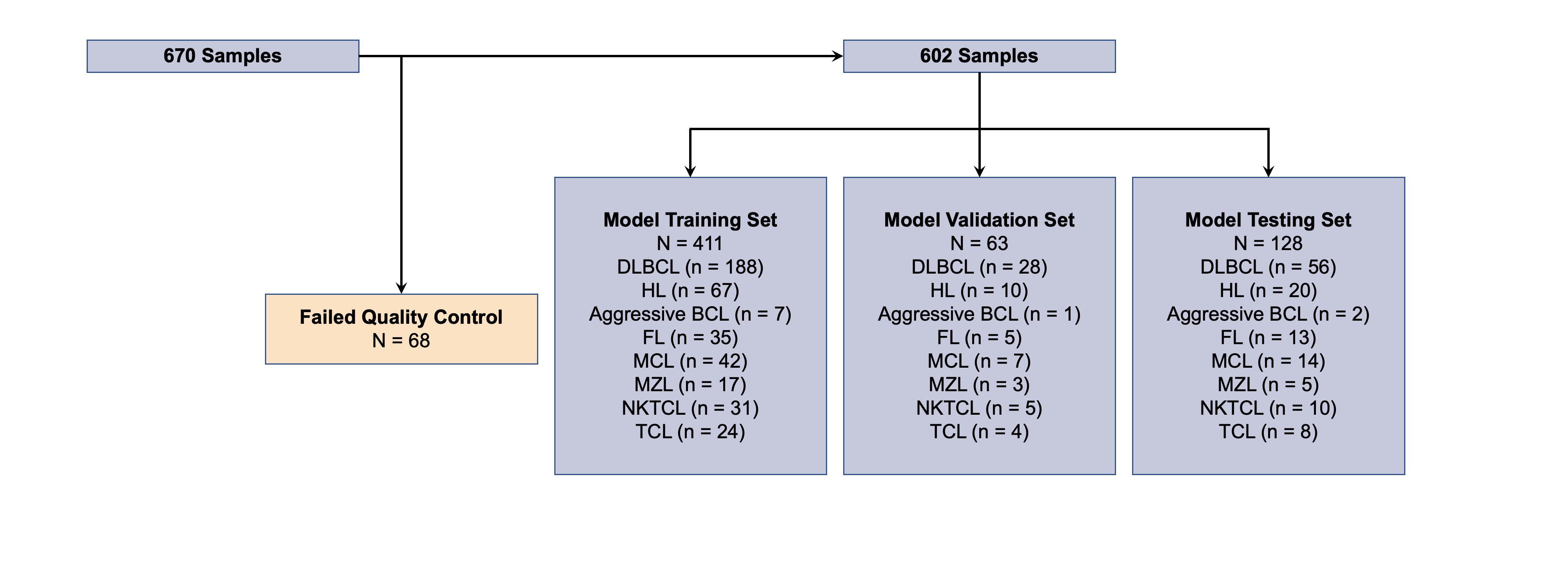}}
\end{figure*}

\begin{figure*}[!ht]
\floatconts
  {fig:SHAPGrouped}
  {\caption{Feature importance analysis by SHapley Additive exPlanation (SHAP) values, while grouping related morphological features into different categories. The nuclear size feature group has the largest mean absolute SHAP value, suggesting that of all nuclear features, size features were the most helpful for classifying DLBCL, CHL, and MCL.}}
  {\includegraphics[width=1.0\linewidth]{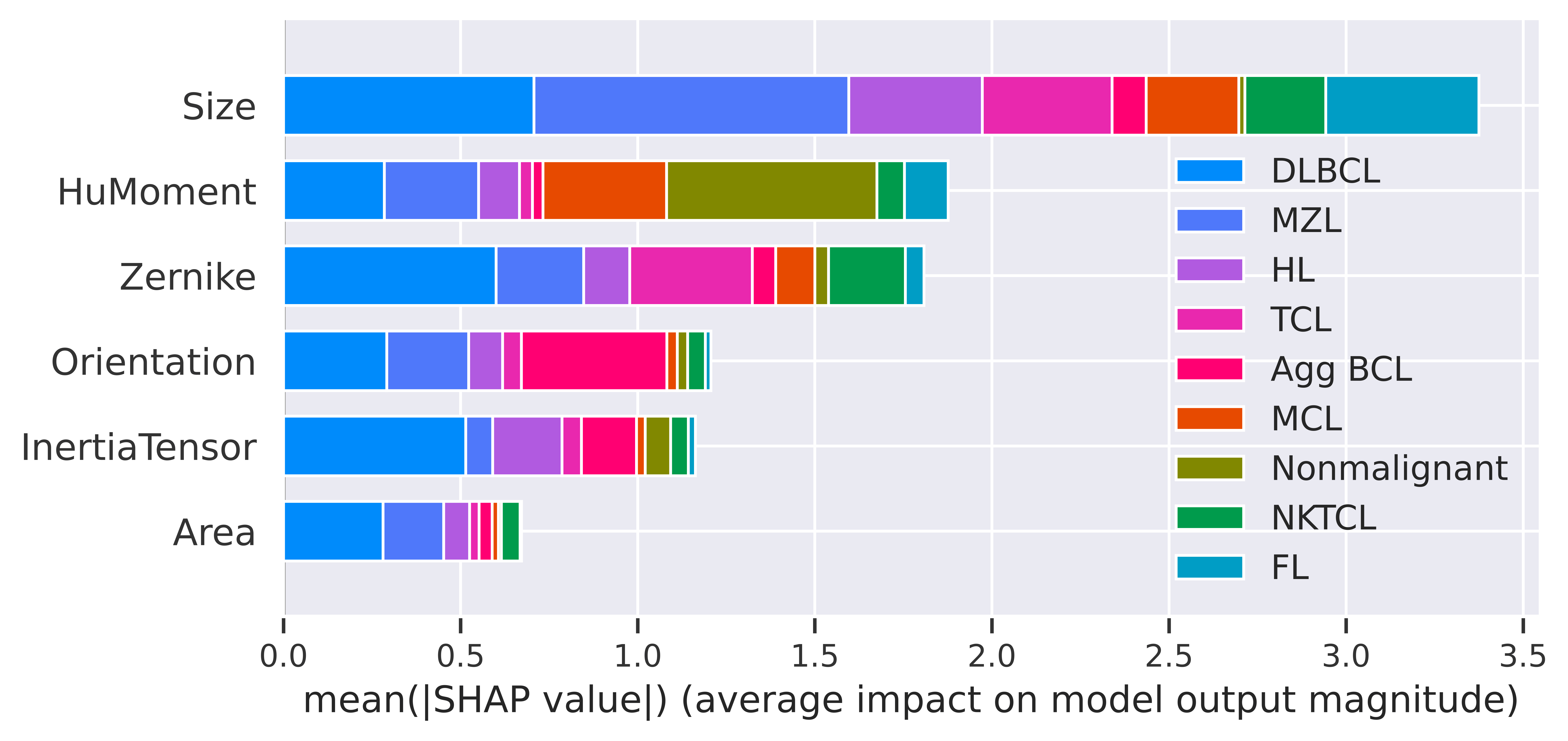}}

\end{figure*}

\begin{figure*}[!ht]
\floatconts
  {fig:ShapTop20}
  {\caption{The SHAP value breakdown of the top 20 morphological features in each diagnosis with the most important features in the “size” group circled.}}
  {\includegraphics[width=1.0\linewidth]{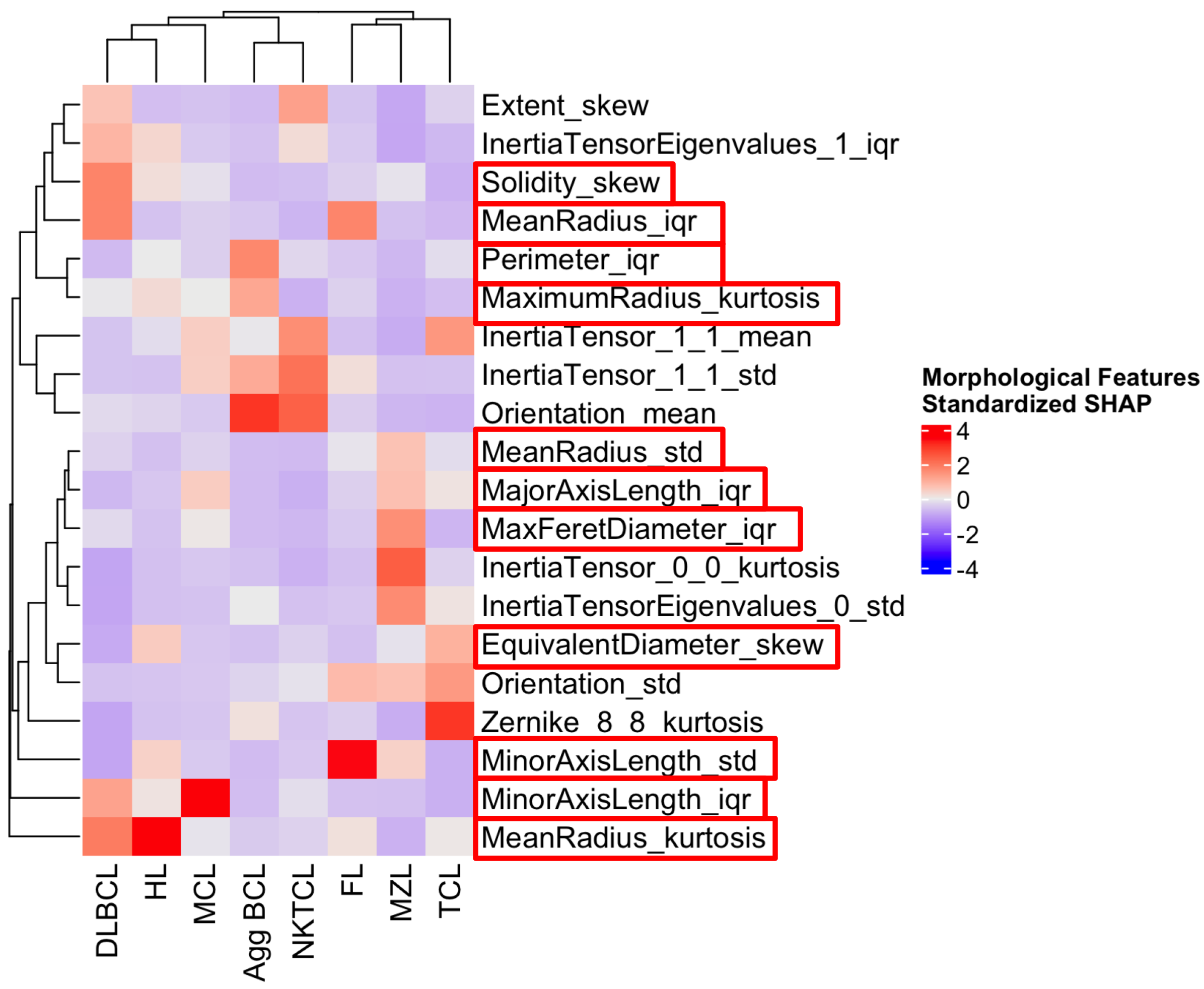}}
\end{figure*}

\clearpage

\counterwithin{table}{section}
\section{Supplementary Tables}\label{apd:second}

\begin{table}[!htbp]
\centering
\begin{tabular}{c | c}
    \hline
    \textbf{Diagnosis} & \textbf{Number of Cases} \\
    \hline
    DLBCL & 272 \\
    CHL & 97 \\
    Agg BCL & 10 \\
    FL & 53 \\
    MCL & 63 \\
    MZL & 25 \\
    NKTCL & 46 \\
    TCL & 36 \\
    \hline
\end{tabular}
\caption{Dataset distribution presenting the number of available cases per lymphoma subtype.}  
\label{table1}
\end{table}

\begin{table}[!htbp]
\centering
\begin{tabular}{c | c | c }
    \hline
    \textbf{Diagnosis} & \textbf{Patches (train split)} & \textbf{Cases (train split)} \\
    \hline
    DLBCL & 32926 & 188 \\
    CHL & 11829 & 67 \\
    Agg BCL & 1303 & 7 \\
    FL & 6562 & 35 \\
    MCL & 7559 & 42 \\
    MZL & 3321 & 17 \\
    NKTCL & 5458 & 31 \\
    TCL & 4198 & 24 \\
    \hline
\end{tabular}
\caption{Dataset size: the number of patches/cases in the training set per-class.}  
\label{table2}
\end{table}

\begin{table*}[!htbp]
\centering
\begin{tabular}{>{\centering\arraybackslash} m{0.15\linewidth}|>{\centering\arraybackslash}m{0.2\linewidth}|>{\centering\arraybackslash}m{0.5\linewidth}}
    \hline
    \textbf{Feature Name} & \textbf{Feature Type} & \textbf{Feature Description} \\
    \hline
    Area & Morphological (Area Shape) & The number of pixels in the region. \\
    \hline
    Bounding Box Area & Morphological (Area Shape) & The area of a box containing the object. \\
    \hline
    Bounding Box Min/Max & Spatial / Architectural & The minimum/maximum x-, y-, and (for 3D objects) z- coordinates of the object. \\
    \hline
    Center\_X, Center\_Y, Center\_Z & Spatial / Architectural & The x-, y-, and (for 3D objects) z- coordinates of the point farthest away from any object edge (the centroid). \\
    \hline
    Central Moment Features & Spatial / Architectural & Similar to spatial moments, but normalized to the object’s centroid. These are therefore not influenced by an object’s location within an image. \\
    \hline
    Compactness & Morphological (Area Shape) & The mean squared distance of the object’s pixels from the centroid divided by the area. A filled circle will have a compactness of 1, with irregular objects or objects with holes having a value greater than 1. \\
    \hline
    Convex Hull Area & Morphological (Area Shape) & The number of pixels in the convex hull of the object. \\
    \hline
    Eccentricity & Morphological (Area Shape) & The eccentricity of the ellipse that has the same second-moments as the region. The eccentricity is the ratio of the distance between the foci of the ellipse and its major axis length. The value is between 0 and 1 (0 and 1 are degenerate cases; an ellipse whose eccentricity is 0 is actually a circle, while an ellipse whose eccentricity is 1 is a line segment.) \\
    \hline
    Equivalent Diameter & Morphological (Area Shape) & The diameter of a circle or sphere with the same area as the object. \\
    \hline
    Euler Number & Morphological (Area Shape) & The number of objects in the region minus the number of holes in those objects, assuming 8-connectivity. \\
    \hline
    Form Factor & Morphological (Area Shape) & Calculated as $4*\pi*{\left(\frac{\text{Area}}{\text{Perimeter}}\right)}^2.$ Equals 1 for a perfectly circular object. \\
    \hline
    Hu Moment Features & Morphological (Area Shape) & Hu’s set of image moment features. These are not altered by the object’s location, size or rotation. This means that they primarily describe the shape of the object. \\
    \hline
    Inertia Tensor Features & Morphological (Area Shape) & A representation of rotational inertia of the object relative to its center. \\
    \hline
    Inertia Tensor Eigenvalues Features & Morphological (Area Shape) & Values describing the movement of the Inertia Tensor array. \\
    \hline
    Integrated Intensity & Texture / Intensity & The sum of the pixel intensities within an object. \\
    \hline
    Integrated Intensity Edge & Texture / Intensity & The sum of the edge pixel intensities of an object. \\
    \hline
    Lower Quartile Intensity & Texture / Intensity & The intensity value of the pixel for which 25\% of the pixels in the object have lower values. \\
    \end{tabular}
\label{table3}
\end{table*}

\begin{table*}[!htbp]
\centering
\begin{tabular}{>{\centering\arraybackslash} m{0.15\linewidth}|>{\centering\arraybackslash}m{0.2\linewidth}|>{\centering\arraybackslash}m{0.5\linewidth}}
    \hline
    \textbf{Feature Name} & \textbf{Feature Type} & \textbf{Feature Description} \\
    \hline
    MAD Intensity & Texture / Intensity & The median absolute deviation (MAD) value of the intensities within the object. The MAD is defined as the $\text{median}(\mid x_i - \text{median}(x) \mid)$. \\
    \hline
    Major/Minor Axis Length & Morphological (Area Shape) & The length (in pixels) of the major/minor axis of the ellipse that has the same normalized second central moments as the region. \\
    \hline
    Mass Displacement & Texture / Intensity & The distance between the centers of gravity in the gray-level representation of the object and the binary representation of the object. \\
    \hline
    Max Intensity & Texture / Intensity & The maximal pixel intensity within an object. \\
    \hline
    Max Intensity Edge & Texture / Intensity & The maximal edge pixel intensity of an object. \\
    \hline
    Maximum Radius & Morphological (Area Shape) & The maximum distance of any pixel in the object to the closest pixel outside of the object. For skinny objects, this is $\frac{1}{2}$ of the maximum width of the object. \\
    \hline
    Max/Min Feret Diameter & Morphological (Area Shape) & The Feret diameter is the distance between two parallel lines tangent on either side of the object (imagine taking a caliper and measuring the object at various angles). The maximum and minimum Feret diameters are the largest and smallest possible diameters, rotating the calipers along all possible angles. \\
    \hline
    Mean Intensity & Texture / Intensity & The average pixel intensity within an object. \\
    \hline
    Mean Intensity Edge & Texture / Intensity & The average edge pixel intensity of an object. \\
    \hline
    Mean Radius & Morphological (Area Shape) & The mean distance of any pixel in the object to the closest pixel outside of the object. \\
    \hline
    Median Intensity & Texture / Intensity & The median intensity value within the object. \\
    \hline
    Median Radius & Morphological (Area Shape) & The median distance of any pixel in the object to the closest pixel outside of the object. \\
    \hline
    Min Intensity & Texture / Intensity & The minimal pixel intensity within an object. \\
    \hline
    Min Intensity Edge & Texture / Intensity & The minimal edge pixel intensity of an object. \\
    \hline
    Normalized Moment Features & Morphological (Area Shape) & Similar to central moments, but further normalized to be scale invariant. These moments are therefore not impacted by an object’s size (or location). \\
    \hline
    Orientation & Morphological (Area Shape) & The angle (in degrees ranging from -90 to 90 degrees) between the x-axis and the major axis of the ellipse that has the same second-moments as the region. \\
    \hline
    Perimeter & Morphological (Area Shape) & The total number of pixels around the boundary of each region in the image. \\
\end{tabular}
\label{table3.1}
\end{table*}

\begin{table*}[!htbp]
\centering
\begin{tabular}{>{\centering\arraybackslash} m{0.15\linewidth}|>{\centering\arraybackslash}m{0.2\linewidth}|>{\centering\arraybackslash}m{0.5\linewidth}}
    \hline
    \textbf{Feature Name} & \textbf{Feature Type} & \textbf{Feature Description} \\
    \hline
    Solidity & Morphological (Area Shape) & The proportion of the pixels in the convex hull that are also in the object, i.e., ObjectArea/ConvexHullArea. \\
    \hline
    Spatial Moment Features & Spatial / Architectural & A series of weighted averages representing the shape, size, rotation and location of the object. \\
    \hline
    Std Intensity & Texture / Intensity & The standard deviation of the pixel intensities within an object. \\
    \hline
    Std Intensity Edge & Texture / Intensity & The standard deviation of the edge pixel intensities of an object. \\
    \hline
    Upper Quartile Intensity & Texture / Intensity & The intensity value of the pixel for which 75\% of the pixels in the object have lower values. \\
    \hline
    Zernike shape Features & Morphological (Area Shape) & These metrics of shape describe a binary object (or more precisely, a patch with background and an object in the center) in a basis of Zernike polynomials, using the coefficients as features (Boland et al., 1998). Currently, Zernike polynomials from order 0 to order 9 are calculated, giving in total 30 measurements. While there is no limit to the order which can be calculated (and indeed you could add more by adjusting the code), the higher order polynomials carry less information. \\
\end{tabular}
\caption{The full list of features extracted by our CellProfiler pipeline. All feature definitions are provided in: \url{https://cellprofiler-manual.s3.amazonaws.com/CellProfiler-4.0.5/modules/measurement.html} \citep{Stirling2021-nr}.}  
\label{table3.2}
\end{table*}

\begin{table*}
\centering
\begin{tabular}{l|c|c|c|c}
\hline
\textbf{Study} &
  \multicolumn{1}{l|}{\textbf{\begin{tabular}[c]{@{}l@{}}Current\\ study\end{tabular}}} &
  \multicolumn{1}{l|}{\textbf{\citet{Zhu2022}}} &
  \multicolumn{1}{l|}{\textbf{\citet{Steinbuss}}} &
  \multicolumn{1}{l}{\textbf{\citet{Mohlman2020-os}}} \\ \hline
 &
  \multicolumn{1}{l|}{\begin{tabular}[c]{@{}l@{}}Large non-\\ Hodgkin B-\\ cell lymphoma\\ (n = 286),\\ Total small\\ B-cell\\ lymphomas\\ (n = 154),\\ CHL (n = 91),\\ Total T and\\ NK cell\\ lymphoma\\ (n = 96), \\ TLL (n = 6),\\ BLL (n = 5),\\ Plasma cell\\ neoplasm\\ (n = 8),\\ Histiocytic\\ sarcoma\\ (n = 1),\\ RFH (n = 23),\\ Total = 670\end{tabular}} &
  \multicolumn{1}{l|}{\begin{tabular}[c]{@{}l@{}}FL\\ (n = 1129),\\ CLL (n = 212),\\ MCL (n = 613),\\ DLBCL\\ (n = 1002),\\ CHL (n = 232),\\ MZL (n = 888),\\ AITL (n = 76),\\ BL (n = 95),\\ Total = 4247\end{tabular}} &
  \multicolumn{1}{l|}{\begin{tabular}[c]{@{}l@{}}CLL (n = 129),\\ DLBCL (n = 119),\\ Control LNs\\ (n = 381),\\ Total = 629\end{tabular}} &
  \multicolumn{1}{l}{\begin{tabular}[c]{@{}l@{}}DLBCL (n = 36),\\ BL (n = 34),\\ Total = 70\end{tabular}} \\ \hline
\begin{tabular}[c]{@{}l@{}}Number of\\ diagnostic\\ categories\end{tabular} &
  8 &
  8 &
  3 &
  2 \\ \hline
\begin{tabular}[c]{@{}l@{}}Average\\ number\\ of cases per\\ diagnostic\\ category\end{tabular} &
  84 &
  531 &
  210 &
  35 \\ \hline
\begin{tabular}[c]{@{}l@{}}Interpret-\\ ability\\ of models\end{tabular} &
  Yes &
  \begin{tabular}[c]{@{}c@{}}Not\\ described\end{tabular} &
  No &
  No \\ \hline
\begin{tabular}[c]{@{}l@{}}Accuracy of\\ best model\end{tabular} &
  85\% &
  81\% &
  95.56\% &
  94\% \\ \hline
\end{tabular}
\label{table4}
\end{table*}

\begin{table*}
\centering
\begin{tabular}{l|c|c|c}
\hline
\textbf{Study} &
  \multicolumn{1}{l|}{\textbf{\citet{Miyoshi2020-lv}}} &
  \multicolumn{1}{l|}{\textbf{\citet{Li2020-dg}}} &
  \multicolumn{1}{l}{\textbf{\citet{Zhang2020-hu}}} \\ \hline
 &
  \begin{tabular}[c]{@{}c@{}}DLBCL (n = 259),\\ FL (n = 89),\\ RFH (n = 40),\\ Total = 388\end{tabular} &
  \begin{tabular}[c]{@{}c@{}}DLBCL\\ (n = 867),\\ Non-DLBCL\\ (n = 887),\\ Total = 1754\end{tabular} &
  \begin{tabular}[c]{@{}c@{}}CLL (n = 113),\\ FL (n = 139),\\ MCL (n = 122),\\ Total = 374\end{tabular} \\ \hline
\begin{tabular}[c]{@{}l@{}}Number of\\ diagnostic\\ categories\end{tabular} &
  3 &
  2 &
  3 \\ \hline
\begin{tabular}[c]{@{}l@{}}Average\\ number\\ of cases per\\ diagnostic\\ category\end{tabular} &
  129 &
  877 &
  125 \\ \hline
\begin{tabular}[c]{@{}l@{}}Interpret-\\ ability\\ of models\end{tabular} &
  No &
  No &
  No \\ \hline
\begin{tabular}[c]{@{}l@{}}Accuracy of\\ best model\end{tabular} &
  97\% &
  99.71-100\% &
  99.2-100\% \\ \hline
\end{tabular}
\label{table4.1}
\end{table*}

\begin{table*}
\centering
\begin{tabular}{l|c|c|c}
\hline
\textbf{Study} &
  \multicolumn{1}{l|}{\textbf{\citet{Achi2019-ik}}} &
  \multicolumn{1}{l|}{\textbf{\citet{Brancati2019-nf}}} &
  \multicolumn{1}{l}{\textbf{\citet{Janowczyk2016-aj}}} \\ \hline
 &
  \begin{tabular}[c]{@{}c@{}}Benign (n = 32),\\ DLBCL (n = 32),\\ BL (n = 32),\\ SLL (n = 32),\\ Total = 128\end{tabular} &
  \begin{tabular}[c]{@{}c@{}}CLL (n = 113),\\ FL (n = 139),\\ MCL (n = 122),\\ Total = 374\end{tabular} &
  \begin{tabular}[c]{@{}c@{}}CLL (n = 113),\\ FL (n = 139),\\ MCL (n = 122),\\ Total = 374\end{tabular} \\ \hline
\begin{tabular}[c]{@{}l@{}}Number of\\ diagnostic\\ categories\end{tabular} &
  4 &
  3 &
  3 \\ \hline
\begin{tabular}[c]{@{}l@{}}Average\\ number\\ of cases per\\ diagnostic\\ category\end{tabular} &
  32 &
  125 &
  125 \\ \hline
\begin{tabular}[c]{@{}l@{}}Interpret-\\ ability\\ of models\end{tabular} &
  No &
  No &
  No \\ \hline
\begin{tabular}[c]{@{}l@{}}Accuracy of\\ best model\end{tabular} &
  95\% &
  97.06\% &
  96.58\% \\ \hline
\end{tabular}
\caption{Summary of case types assessed in studies that apply computer vision tools to lymphoma diagnosis. CLL = chronic lymphocytic leukemia, SLL = small lymphocytic lymphoma, FL = follicular lymphoma, MCL = mantle cell lymphoma, DLBCL = diffuse large B-cell lymphoma, BL = Burkitt lymphoma, RFH = reactive follicular hyperplasia, TLL = T-lymphoblastic lymphoma, BLL = B-lymphoblastic leukemia, LN = lymph nodes}
\label{table4.2}
\end{table*}

\begin{table*}
\begin{center}
\begin{tabular}{l|c|c|c|c}
\hline
\multicolumn{1}{c|}{\textbf{Study}} &
  \textbf{\begin{tabular}[c]{@{}c@{}}Current\\ study\end{tabular}} &
  \textbf{\begin{tabular}[c]{@{}c@{}}\citet{Yu2021} \end{tabular}} &
  \textbf{\begin{tabular}[c]{@{}c@{}}\citet{Gupta2010-ff}\end{tabular}} &
  \textbf{\begin{tabular}[c]{@{}c@{}} \citet{Lesty1986-ze} \end{tabular}} \\ \hline
 &
  \multicolumn{1}{l|}{\begin{tabular}[c]{@{}l@{}}Large non-\\ Hodgkin B-\\ cell lymphoma\\ (n = 286),\\ Total small\\ B-cell\\ lymphomas\\ (n = 154), \\ CHL \\ (n = 91), \\ Total T and\\ NK cell\\ lymphoma\\ (n = 96),\\ TLL\\ (n = 6),\\ BLL\\ (n = 5),\\ Plasma cell\\ neoplasm\\ (n = 8),\\ Histiocytic\\ sarcoma\\ (n = 1),\\ RFH\\ (n = 23),\\ Total = 670\end{tabular}} &
  \multicolumn{1}{l|}{\begin{tabular}[c]{@{}l@{}}MEITL (n = 26),\\ ITL-NOS \\ (n = 10),\\ Borderline cases \\ (n = 4),\\ Total = 40\end{tabular}} &
  \multicolumn{1}{l|}{\begin{tabular}[c]{@{}l@{}}Fine needle \\ aspiration cytology \\ smears of 5 cases. \\ SLL (n = 13), \\ FL (n = 9), \\ DLBCL centroblastic \\ (n = 14), DLBCL \\ anaplastic (n = 4), \\ DLBCL immunoblastic \\ (n = 2), lymphoblastic \\ lymphoma (n = 8),  \\ Total = 50 \end{tabular}} &
  \multicolumn{1}{l}{\begin{tabular}[c]{@{}l@{}}Lymphomas\\ classified according \\ to the International\\ Working Formulation.\\ \\ Small noncleaved\\ lymphocytes,\\ CLL (n = 8),\\ Predominantly small \\ cleaved cell; follicular\\ or diffuse or both \\ (n = 7), Mixed diffuse\\ small and large; \\ cleaved or noncleaved\\ cells (n = 13), Large \\ diffuse noncleaved\\ cells (n = 7), Large \\ diffuse cleaved cells\\ (n = 10), \\ Total = 45\end{tabular}} \\ \hline
\begin{tabular}[c]{@{}l@{}}Total number\\ of cases\end{tabular} &
  670 &
  40 &
  50 &
  45 \\ \hline
\begin{tabular}[c]{@{}l@{}}Number of \\ diagnostic\\ categories\end{tabular} &
  8 &
  2 &
  3* &
  5** \\ \hline
\begin{tabular}[c]{@{}l@{}}Average number \\ of cases per \\ diagnostic \\ category\end{tabular} &
  84 &
  20 &
  16 &
  9 \\ \hline
\begin{tabular}[c]{@{}l@{}}Interpretability \\ of models\end{tabular} &
  Yes &
  Yes &
  Yes &
  Yes \\ \hline
\begin{tabular}[c]{@{}l@{}}Accuracy of\\ best model\end{tabular} &
  85\% &
  95\% &
  97\% &
  97\% \\ \hline
\end{tabular}
\caption{Notable features of studies that identify interpretable features for lymphoma diagnosis. CLL = chronic lymphocytic leukemia, SLL = small lymphocytic lymphoma, FL = follicular lymphoma, MCL = mantle cell lymphoma, DLBCL = diffuse large B-cell lymphoma, BL = Burkitt lymphoma, RFH = reactive follicular hyperplasia, TLL = T-lymphoblastic lymphoma, BLL = B-lymphoblastic leukemia, WSI = whole slide images, TMA = tissue microarrays, MEITL = monomorphic epitheliotropic intestinal T-cell lymphoma, ITL-NOS = intestinal T-cell lymphoma, not otherwise specified.}
\label{table5}
\end{center}
\footnotesize{$^*$By the World Health Organization Classification, centroblastic, immunoblastic, and anaplastic variants of diffuse large B-cell lymphoma are not separate diagnostic categories.}\\
\footnotesize{$^{**}$Diagnostic categories were based on the International Working Formulation.}\\
\end{table*}

\begin{table*}[!htbp]
\centering

\begin{tabular}{c|c|c|c|c|c}
\hline
\multirow{2}{*}{\textbf{Model Type}} &
  \multirow{2}{*}{\textbf{Model Features}} &
  \multirow{2}{*}{\textbf{\begin{tabular}[c]{@{}c@{}}\# of\\ features\end{tabular}}} &
  \multirow{2}{*}{\textbf{\begin{tabular}[c]{@{}c@{}}Test\\ Accuracy\end{tabular}}} &
  \multirow{2}{*}{\textbf{\begin{tabular}[c]{@{}c@{}}Test\\ Sensitivity\end{tabular}}} &
  \multirow{2}{*}{\textbf{\begin{tabular}[c]{@{}c@{}}Test\\ Specificity\end{tabular}}} \\
 &
   &
   &
   &
   &
   \\ \hline
\multirow{6}{*}{\textbf{\begin{tabular}[c]{@{}c@{}}Nuclear\\ Morphology\end{tabular}}} &
  \begin{tabular}[c]{@{}c@{}}Nuclear\\ Morphological\\ Features\end{tabular} &
  310 &
  $59.7 \pm 8.5$ &
  $58.9 \pm 10.1$ &
  $84.9 \pm 4.2$ \\ \cline{2-6} 
 &
  \begin{tabular}[c]{@{}c@{}}Nuclear\\ Intensity\\ Features\end{tabular} &
  225 &
  $58.9 \pm 7.8$ &
  $57.4 \pm 9.3$ &
  $84.7 \pm 4.4$ \\ \cline{2-6} 
 &
  \begin{tabular}[c]{@{}c@{}}Cytoplasmic\\ Features\end{tabular} &
  470 &
  $57.4 \pm 9.3$ &
  $55.0 \pm 9.3$ &
  $84.2 \pm 4.4$ \\ \cline{2-6} 
 &
  \begin{tabular}[c]{@{}c@{}}Nuclear\\ Morphological +\\ Intensity\\ Features\end{tabular} &
  630 &
  $61.2 \pm 8.6$ &
  $60.5 \pm 10.0$ &
  $87.1 \pm 4.2$ \\ \cline{2-6} 
 &
  \begin{tabular}[c]{@{}c@{}}Nuclear +\\ Cytoplasmic\\ Features\end{tabular} &
  950 &
  $62.0 \pm 8.5$ &
  $59.7 \pm 9.3$ &
  $86.8 \pm 3.7$ \\ \cline{2-6} 
 &
  \begin{tabular}[c]{@{}c@{}}Nuclear\\ Morphological\\ Model:\\ Parsimonious\end{tabular} &
  36 &
  $61.2 \pm 7.8$ &
  $57.4 \pm 9.3$ &
  $84.0 \pm 4.4$ \\ \hline
\multirow{6}{*}{\textbf{\begin{tabular}[c]{@{}c@{}}Architectural\\ Features\end{tabular}}} &
  \begin{tabular}[c]{@{}c@{}}CPArch +\\ CT\end{tabular} &
  217 &
  $44.9 \pm 8.6$ &
  $43.3 \pm 10.6$ &
  $74.7 \pm 5.5$ \\ \cline{2-6} 
 &
  \begin{tabular}[c]{@{}c@{}}Nuclear +\\ CPArch\end{tabular} &
  475 &
  $60.5 \pm 7.7$ &
  $62.2 \pm 9.1$ &
  $86.2 \pm 4.1$ \\ \cline{2-6} 
 &
  \begin{tabular}[c]{@{}c@{}}Nuclear +\\ CPArch +\\ CT\end{tabular} &
  675 &
  $59.8 \pm 7.9$ &
  $52.0 \pm 10.5$ &
  $75.3 \pm 6.2$ \\ \cline{2-6} 
 &
  \begin{tabular}[c]{@{}c@{}}Nuclear +\\ Cytoplasm +\\ Intensity\end{tabular} &
  1530 &
  $62.0 \pm 8.5$ &
  $62.8 \pm 10.1$ &
  $88.8 \pm 3.3$ \\ \cline{2-6} 
 &
  \begin{tabular}[c]{@{}c@{}}Nuclear +\\ Cytoplasm +\\ Intensity +\\ CPArch\end{tabular} &
  1595 &
  $64.3 \pm 8.6$ &
  $66.9 \pm 6.0$ &
  $88.7 \pm 2.6$ \\ \cline{2-6} 
 &
  \begin{tabular}[c]{@{}c@{}}Nuclear +\\ CPArch +\\ Cytoplasm +\\ Intensity +\\ CT\end{tabular} &
  1695 &
  $64.1 \pm 7.8$ &
  $64.1 \pm 8.6$ &
  $86.2 \pm 4.2$ \\ \hline
\multirow{2}{*}{\textbf{\begin{tabular}[c]{@{}c@{}}Deep\\ Learning\end{tabular}}} &
  \begin{tabular}[c]{@{}c@{}}ResNet-50\\ (Self-Supervised\\ H\&E)\end{tabular} &
  N/A &
  $53.5 \pm 8.7$ &
  $55.5 \pm 10.6$ &
  $82.6 \pm 5.3$ \\ \cline{2-6} 
 &
  \begin{tabular}[c]{@{}c@{}}TripletNet\\ finetuned\\ (Camelyon)\end{tabular} &
  N/A &
  $52.8 \pm 8.6$ &
  $53.7 \pm 9.3$ &
  $75.7 \pm 6.0$ \\ \hline
\end{tabular}
\label{table6}
\end{table*}

\begin{table*}
\centering
\begin{tabular}{c|c|c|c}
\hline
\multirow{2}{*}{\textbf{Model Type}} &
  \multirow{2}{*}{\textbf{Model Features}} &
  \multirow{2}{*}{\textbf{\begin{tabular}[c]{@{}c@{}}Test\\ AUROC\end{tabular}}} &
  \multirow{2}{*}{\textbf{\begin{tabular}[c]{@{}c@{}}Test\\ F1 Score\end{tabular}}} \\
 &
   &
   &
   \\ \hline
\multirow{6}{*}{\textbf{\begin{tabular}[c]{@{}c@{}}Nuclear\\ Morphology\end{tabular}}} &
  \begin{tabular}[c]{@{}c@{}}Nuclear\\ Morphological\\ Features\end{tabular} &
  $82.0 \pm 6.3$ &
  $54.1 \pm 8.9$ \\ \cline{2-4} 
 &
  \begin{tabular}[c]{@{}c@{}}Nuclear\\ Intensity\\ Features\end{tabular} &
  $82.8 \pm 5.4$ &
  $54.1 \pm 9.9$ \\ \cline{2-4} 
 &
  \begin{tabular}[c]{@{}c@{}}Cytoplasmic\\ Features\end{tabular} &
  $83.1 \pm 5.9$ &
  $52.8 \pm 9.4$ \\ \cline{2-4} 
 &
  \begin{tabular}[c]{@{}c@{}}Nuclear\\ Morphological +\\ Intensity\\ Features\end{tabular} &
  $84.2 \pm 5.5$ &
  $56.4 \pm 9.3$ \\ \cline{2-4} 
 &
  \begin{tabular}[c]{@{}c@{}}Nuclear +\\ Cytoplasmic\\ Features\end{tabular} &
  $84.6 \pm 6.0$ &
  $56.4 \pm 9.2$ \\ \cline{2-4} 
 &
  \begin{tabular}[c]{@{}c@{}}Nuclear\\ Morphological\\ Model:\\ Parsimonious\end{tabular} &
  $81.5 \pm 6.6$ &
  $57.0 \pm 8.6$ \\ \hline
\multirow{6}{*}{\textbf{\begin{tabular}[c]{@{}c@{}}Architectural\\ Features\end{tabular}}} &
  \begin{tabular}[c]{@{}c@{}}CPArch +\\ CT\end{tabular} &
  $66.5 \pm 7.9$ &
  $39.8 \pm 9.2$ \\ \cline{2-4} 
 &
  \begin{tabular}[c]{@{}c@{}}Nuclear +\\ CPArch\end{tabular} &
  $82.7 \pm 6.7$ &
  $54.7 \pm 9.5$ \\ \cline{2-4} 
 &
  \begin{tabular}[c]{@{}c@{}}Nuclear +\\ CPArch +\\ CT\end{tabular} &
  $76.1 \pm 6.9$ &
  $51.1 \pm 9.4$ \\ \cline{2-4} 
 &
  \begin{tabular}[c]{@{}c@{}}Nuclear +\\ Cytoplasm +\\ Intensity\end{tabular} &
  $85.3 \pm 5.1$ &
  $56.5 \pm 9.0$ \\ \cline{2-4} 
 &
  \begin{tabular}[c]{@{}c@{}}Nuclear +\\ Cytoplasm +\\ Intensity +\\ CPArch\end{tabular} &
  $85.9 \pm 2.9$ &
  $58.5 \pm 9.5$ \\ \cline{2-4} 
 &
  \begin{tabular}[c]{@{}c@{}}Nuclear +\\ CPArch +\\ Cytoplasm +\\ Intensity +\\ CT\end{tabular} &
  $85.5 \pm 5.0$ &
  $58.0 \pm 10.1$ \\ \hline
\multirow{2}{*}{\textbf{\begin{tabular}[c]{@{}c@{}}Deep\\ Learning\end{tabular}}} &
  \begin{tabular}[c]{@{}c@{}}ResNet-50\\ (Self-Supervised\\ H\&E)\end{tabular} &
  $82.4 \pm 3.5$ &
  $51.7 \pm 9.8$ \\ \cline{2-4} 
 &
  \begin{tabular}[c]{@{}c@{}}TripletNet\\ finetuned\\ (Camelyon)\end{tabular} &
  $81.7 \pm 2.3$ &
  $45.7 \pm 9.6$ \\ \hline
\end{tabular}
\caption{Overall performance summary of feature-based models using different feature combinations and deep-learning models. All features are extracted from H\&E stains only. All metrics are calculated using a weighted average across all labels. We weight by the support, the number of true instances for each label. CPArch = CellProfiler Architectural features, CT = clustering tendency.}
\label{table6.1}
\end{table*}

\begin{table*}
\centering
\begin{tabular}{c|l|cccc}
\hline
\multirow{2}{*}{\textbf{Model Type}} &
  \multicolumn{1}{c|}{\multirow{2}{*}{\textbf{Model Features}}} &
  \multicolumn{4}{c}{\textbf{Per-Class Test F1}} \\ \cline{3-6} 
 &
  \multicolumn{1}{c|}{} &
  \multicolumn{1}{c|}{\textbf{DLBCL}} &
  \multicolumn{1}{c|}{\textbf{HL}} &
  \multicolumn{1}{c|}{\textbf{\begin{tabular}[c]{@{}c@{}}Agg\\ BCL\end{tabular}}} &
  \multicolumn{1}{c}{\textbf{FL}} \\ \hline
\multirow{6}{*}{\textbf{\begin{tabular}[c]{@{}c@{}}Nuclear\\ Morphology\end{tabular}}} &
  \begin{tabular}[c]{@{}l@{}}Nuclear\\ Morphological\\ Features\end{tabular} &
  \multicolumn{1}{c|}{$76.2 \pm 8.0$} &
  \multicolumn{1}{c|}{$65.3 \pm 17.3$} &
  \multicolumn{1}{c|}{0.0} &
  \multicolumn{1}{c}{$11.1 \pm 11.1$} \\ \cline{2-6} 
 &
  \begin{tabular}[c]{@{}l@{}}Nuclear\\ Intensity\\ Features\end{tabular} &
  \multicolumn{1}{c|}{$76.0 \pm 6.6$} &
  \multicolumn{1}{c|}{$64.0 \pm 15.1$} &
  \multicolumn{1}{c|}{0.0} &
  \multicolumn{1}{c}{$27.0 \pm 25.6$} \\ \cline{2-6} 
 &
  \begin{tabular}[c]{@{}l@{}}Cytoplasmic\\ Features\end{tabular} &
  \multicolumn{1}{c|}{$77.0 \pm 8.5$} &
  \multicolumn{1}{c|}{$60.0 \pm 14.4$} &
  \multicolumn{1}{c|}{0.0} &
  \multicolumn{1}{c}{$25.0 \pm 22.6$} \\ \cline{2-6} 
 &
  \begin{tabular}[c]{@{}l@{}}Nuclear\\ Morphological + \\ Intensity\\ Features\end{tabular} &
  \multicolumn{1}{c|}{$77.0 \pm 8.0$} &
  \multicolumn{1}{c|}{$69.4 \pm 13.3$} &
  \multicolumn{1}{c|}{0.0} &
  \multicolumn{1}{c}{$28.6 \pm 25.2$} \\ \cline{2-6} 
 &
  \begin{tabular}[c]{@{}l@{}}Nuclear +\\ Cytoplasmic\\ Features\end{tabular} &
  \multicolumn{1}{c|}{$79.0 \pm 6.9$} &
  \multicolumn{1}{c|}{$76.6 \pm 13.6$} &
  \multicolumn{1}{c|}{0.0} &
  \multicolumn{1}{c}{$34.8 \pm 22.3$} \\ \cline{2-6} 
 &
  \begin{tabular}[c]{@{}l@{}}Nuclear\\ Morphological\\ Model:\\ Parsimonious\end{tabular} &
  \multicolumn{1}{c|}{$74.6 \pm 8.0$} &
  \multicolumn{1}{c|}{$69.2 \pm 13.6$} &
  \multicolumn{1}{c|}{0.0} &
  \multicolumn{1}{c}{$38.1 \pm 25.9$} \\ \hline
\multirow{6}{*}{\textbf{\begin{tabular}[c]{@{}c@{}}Architectural\\ Features\end{tabular}}} &
  \begin{tabular}[c]{@{}l@{}}CPArch + \\ CT\end{tabular} &
  \multicolumn{1}{c|}{$68.2 \pm 7.8$} &
  \multicolumn{1}{c|}{$33.3 \pm 17.5$} &
  \multicolumn{1}{c|}{0.0} &
  \multicolumn{1}{c}{$20.0 \pm 20.0$} \\ \cline{2-6} 
 &
  \begin{tabular}[c]{@{}l@{}}Nuclear + \\ CPArch\end{tabular} &
  \multicolumn{1}{c|}{$76.2 \pm 7.0$} &
  \multicolumn{1}{c|}{$69.4 \pm 14.6$} &
  \multicolumn{1}{c|}{0.0} &
  \multicolumn{1}{c}{$11.1 \pm 11.1$} \\ \cline{2-6} 
 &
  \begin{tabular}[c]{@{}l@{}}Nuclear + \\ CPArch + \\ CT\end{tabular} &
  \multicolumn{1}{c|}{$75.8 \pm 7.7$} &
  \multicolumn{1}{c|}{$75.0 \pm 13.4$} &
  \multicolumn{1}{c|}{0.0} &
  \multicolumn{1}{c}{0.0} \\ \cline{2-6} 
 &
  \begin{tabular}[c]{@{}l@{}}Nuclear + \\ Cytoplasm + \\ Intensity\end{tabular} &
  \multicolumn{1}{c|}{$78.1 \pm 7.0$} &
  \multicolumn{1}{c|}{$70.6 \pm 14.5$} &
  \multicolumn{1}{c|}{0.0} &
  \multicolumn{1}{c}{$30.0 \pm 25.2$} \\ \cline{2-6} 
 &
  \begin{tabular}[c]{@{}l@{}}Nuclear + \\ Cytoplasm + \\ Intensity + \\ CPArch\end{tabular} &
  \multicolumn{1}{c|}{$78.7 \pm 7.7$} &
  \multicolumn{1}{c|}{$74.5 \pm 13.4$} &
  \multicolumn{1}{c|}{0.0} &
  \multicolumn{1}{c}{$31.6 \pm 24.8$} \\ \cline{2-6} 
 &
  \begin{tabular}[c]{@{}l@{}}Nuclear + \\ CPArch + \\ Cytoplasm + \\ Intensity + \\ CT\end{tabular} &
  \multicolumn{1}{c|}{$80.0 \pm 7.7$} &
  \multicolumn{1}{c|}{$70.8 \pm 14.4$} &
  \multicolumn{1}{c|}{0.0} &
  \multicolumn{1}{c}{$22.2 \pm 22.2$} \\ \hline
\multirow{2}{*}{\textbf{\begin{tabular}[c]{@{}c@{}}Deep\\ Learning\end{tabular}}} &
  \begin{tabular}[c]{@{}l@{}}ResNet-50\\ (Self-Supervised\\ H\&E)\end{tabular} &
  \multicolumn{1}{c|}{$72.5 \pm 8.2$} &
  \multicolumn{1}{c|}{$59.7 \pm 17.7$} &
  \multicolumn{1}{c|}{0.0} &
  \multicolumn{1}{c}{$27.7 \pm 24.5$} \\ \cline{2-6} 
 &
  \begin{tabular}[c]{@{}l@{}}TripletNet\\ finetuned\\ (Camelyon)\end{tabular} &
  \multicolumn{1}{c|}{$70.3 \pm 8.1$} &
  \multicolumn{1}{c|}{$30.1 \pm 21.0$} &
  \multicolumn{1}{c|}{0.0} &
  \multicolumn{1}{c}{$30.6 \pm 27.7$} \\ \hline
\end{tabular}
\label{table6.2}
\end{table*}

\begin{table*}
\centering
\begin{tabular}{c|l|cccc}
\hline
\multirow{2}{*}{\textbf{Model Type}} &
  \multicolumn{1}{c|}{\multirow{2}{*}{\textbf{Model Features}}} &
  \multicolumn{4}{c}{\textbf{Per-Class Test F1}} \\ \cline{3-6} 
 &
  \multicolumn{1}{c|}{} &
  \multicolumn{1}{c|}{\textbf{MCL}} &
  \multicolumn{1}{c|}{\textbf{MZL}} &
  \multicolumn{1}{c|}{\textbf{NKTCL}} &
  \textbf{TCL} \\ \hline
\multirow{6}{*}{\textbf{\begin{tabular}[c]{@{}c@{}}Nuclear\\ Morphology\end{tabular}}} &
  \begin{tabular}[c]{@{}l@{}}Nuclear\\ Morphological\\ Features\end{tabular} &
  \multicolumn{1}{c|}{$51.6 \pm 19.4$} &
  \multicolumn{1}{c|}{$22.2 \pm 22.2$} &
  \multicolumn{1}{c|}{$42.9 \pm 29.1$} &
  0.0 \\ \cline{2-6} 
 &
  \begin{tabular}[c]{@{}l@{}}Nuclear\\ Intensity\\ Features\end{tabular} &
  \multicolumn{1}{c|}{$56.0 \pm 20.2$} &
  \multicolumn{1}{c|}{0.0} &
  \multicolumn{1}{c|}{$29.0 \pm 29.0$} &
  0.0 \\ \cline{2-6} 
 &
  \begin{tabular}[c]{@{}l@{}}Cytoplasmic\\ Features\end{tabular} &
  \multicolumn{1}{c|}{$51.0 \pm 19.6$} &
  \multicolumn{1}{c|}{0.0} &
  \multicolumn{1}{c|}{$25.0 \pm 25.0$} &
  0.0 \\ \cline{2-6} 
 &
  \begin{tabular}[c]{@{}l@{}}Nuclear\\ Morphological + \\ Intensity\\ Features\end{tabular} &
  \multicolumn{1}{c|}{$66.7 \pm 16.6$} &
  \multicolumn{1}{c|}{0.0} &
  \multicolumn{1}{c|}{$26.7 \pm 26.7$} &
  0.0 \\ \cline{2-6} 
 &
  \begin{tabular}[c]{@{}l@{}}Nuclear +\\ Cytoplasmic\\ Features\end{tabular} &
  \multicolumn{1}{c|}{$51.6 \pm 19.4$} &
  \multicolumn{1}{c|}{0.0} &
  \multicolumn{1}{c|}{$14.3 \pm 14.3$} &
  0.0 \\ \cline{2-6} 
 &
  \begin{tabular}[c]{@{}l@{}}Nuclear\\ Morphological\\ Model:\\ Parsimonious\end{tabular} &
  \multicolumn{1}{c|}{$64.5 \pm 17.9$} &
  \multicolumn{1}{c|}{$25.0 \pm 25.0$} &
  \multicolumn{1}{c|}{$26.7 \pm 26.7$} &
  0.0 \\ \hline
\multirow{6}{*}{\textbf{\begin{tabular}[c]{@{}c@{}}Architectural\\ Features\end{tabular}}} &
  \begin{tabular}[c]{@{}l@{}}CPArch + \\ CT\end{tabular} &
  \multicolumn{1}{c|}{$16.7 \pm 16.7$} &
  \multicolumn{1}{c|}{$33.3 \pm 33.3$} &
  \multicolumn{1}{c|}{0.0} &
  0.0 \\ \cline{2-6} 
 &
  \begin{tabular}[c]{@{}l@{}}Nuclear + \\ CPArch\end{tabular} &
  \multicolumn{1}{c|}{$53.3 \pm 18.9$} &
  \multicolumn{1}{c|}{$22.2 \pm 22.2$} &
  \multicolumn{1}{c|}{$40.0 \pm 30.0$} &
  0.0 \\ \cline{2-6} 
 &
  \begin{tabular}[c]{@{}l@{}}Nuclear + \\ CPArch + \\ CT\end{tabular} &
  \multicolumn{1}{c|}{$46.7 \pm 20.0$} &
  \multicolumn{1}{c|}{0.0} &
  \multicolumn{1}{c|}{$16.7 \pm 16.7$} &
  0.0 \\ \cline{2-6} 
 &
  \begin{tabular}[c]{@{}l@{}}Nuclear + \\ Cytoplasm + \\ Intensity\end{tabular} &
  \multicolumn{1}{c|}{$62.1 \pm 17.9$} &
  \multicolumn{1}{c|}{0.0} &
  \multicolumn{1}{c|}{$25.0 \pm 25.0$} &
  0.0 \\ \cline{2-6} 
 &
  \begin{tabular}[c]{@{}l@{}}Nuclear + \\ Cytoplasm + \\ Intensity + \\ CPArch\end{tabular} &
  \multicolumn{1}{c|}{$71.0 \pm 16.0$} &
  \multicolumn{1}{c|}{0.0} &
  \multicolumn{1}{c|}{$25.0 \pm 25.0$} &
  0.0 \\ \cline{2-6} 
 &
  \begin{tabular}[c]{@{}l@{}}Nuclear + \\ CPArch + \\ Cytoplasm + \\ Intensity + \\ CT\end{tabular} &
  \multicolumn{1}{c|}{$71.0 \pm 16.0$} &
  \multicolumn{1}{c|}{0.0} &
  \multicolumn{1}{c|}{$26.7 \pm 26.7$} &
  0.0 \\ \hline
\multirow{2}{*}{\textbf{\begin{tabular}[c]{@{}c@{}}Deep\\ Learning\end{tabular}}} &
  \begin{tabular}[c]{@{}l@{}}ResNet-50\\ (Self-Supervised\\ H\&E)\end{tabular} &
  \multicolumn{1}{c|}{$51.1 \pm 23.9$} &
  \multicolumn{1}{c|}{0.0} &
  \multicolumn{1}{c|}{$34.1 \pm 25.2$} &
  0.0 \\ \cline{2-6} 
 &
  \begin{tabular}[c]{@{}l@{}}TripletNet\\ finetuned\\ (Camelyon)\end{tabular} &
  \multicolumn{1}{c|}{$49.2 \pm 20.6$} &
  \multicolumn{1}{c|}{0.0} &
  \multicolumn{1}{c|}{$29.0 \pm 29.0$} &
  0.0 \\ \hline
\end{tabular}
\caption{Per-class performance summary of feature-based models using different feature combinations and deep-learning models. All features are extracted from H\&E stains only. CPArch = CellProfiler Architectural features, CT = clustering tendency.} 
\label{table6.3}
\end{table*}

\begin{table*}[!htbp]
\centering
\begin{tabular}{>{\centering\arraybackslash} m{0.25\linewidth}|>{\centering\arraybackslash}m{0.12\linewidth}|>{\centering\arraybackslash}m{0.12\linewidth}|>{\centering\arraybackslash}m{0.12\linewidth}|>{\centering\arraybackslash}m{0.12\linewidth}|>{\centering\arraybackslash}m{0.12\linewidth}}
    \hline
    \textbf{Method} & \textbf{Test Accuracy} & \textbf{Test Weighted Sensitivity}  & \textbf{Test Weighted Specificity} & \textbf{Test Weighted AUROC} & \textbf{F1 Score} \\
    \hline
    Baseline: Hematopathologist 1 on H\&E TMAs & $73.0 \pm 7.4$ & $72.5 \pm 8.2$ & $73.2 \pm 7.6$ & N/A & $73.1 \pm 6.6$ \\
    \hline
    Baseline: Hematopathologist 2 on H\&E TMAs & $73.0  \pm 6.7$ & $72.9 \pm 8.1$ & $73.0 \pm 7.9$ & N/A & $73.0 \pm 7.0$ \\
    \hline
    Baseline: Hematopathologist 3 on WSIs & $83.8 \pm 5.4$ & $82.8 \pm 6.2$ & $82.6 \pm 6.6$ & N/A & $83.8 \pm 5.4$ \\
    \hline
    Baseline: General Pathologist on WSIs & $73.6 \pm 6.8$ & $69.0 \pm 7.6$ & $68.3 \pm 8.6$ & N/A & $73.1 \pm 6.9$ \\
    \hline
    Best Model with nuclear size/area features only & $76.0 \pm 6.9$ & $76.0 \pm 8.5$ & $74.7 \pm 9.0$ & $81.7 \pm 7.5$ & $75.9 \pm 7.1$ \\
    \hline
    Best H\&E Model & $79.8 \pm 6.2$ & $81.3 \pm 7.1$ & $81.8 \pm 6.8$ & $78.9 \pm 6.9$ & $81.5 \pm 6.9$ \\
    \hline
\end{tabular}
\caption{Performance comparison of the Best H\&E Model to hematopathologists on the DLBCL vs non-DLBCL classification task. }  
\label{table7}
\end{table*}

\begin{table*}[!htbp]
\centering
\begin{tabular}{>{\centering\arraybackslash} m{0.25\linewidth}|>{\centering\arraybackslash}m{0.12\linewidth}|>{\centering\arraybackslash}m{0.12\linewidth}|>{\centering\arraybackslash}m{0.12\linewidth}|>{\centering\arraybackslash}m{0.12\linewidth}|>{\centering\arraybackslash}m{0.12\linewidth}}
    \hline
    \textbf{Method} & \textbf{Test Accuracy} & \textbf{Test Weighted Sensitivity}  & \textbf{Test Weighted Specificity} & \textbf{Test Weighted AUROC} & \textbf{Test Weighted F1 Score} \\
    \hline
    Baseline: Best H\&E Model & $69.0 \pm 7.7$ & $69.8 \pm 8.5$ & $86.4 \pm 4.5$ & $85.5 \pm 5.2$ & $65.7 \pm 8.4$ \\
    \hline
    Baseline: CD10, CD20, CD3, EBV ISH, BCL1, CD30 & $75.2 \pm 7.0$ & $75.2 \pm 7.0$ & $86.4 \pm 5.4$ & $89.7 \pm 4.5$ & $70.9 \pm 8.7$ \\
    \hline
    Baseline: best model with all immunostains & $86.1 \pm 6.1$ & $86.0 \pm 7.0$ & $93.2 \pm 3.4$ & $96.7 \pm 2.2$ & $85.1 \pm 6.0$ \\
    \hline
    H\&E + CD20 & $78.3 \pm 6.2$ & $79.1 \pm 7.7$ & $93.1 \pm 3.5$ & $93.3 \pm 3.5$ & $78.1 \pm 6.9$ \\
    \hline
    H\&E + CD3 & $79.1 \pm 6.2$ & $81.4 \pm 7.0$ & $92.9 \pm 3.8$ & $93.2 \pm 3.3$ & $78.3 \pm 7.4$ \\
    \hline
    H\&E + CD3, CD20 & $82.9 \pm 6.2$ & $82.9 \pm 7.8$ & $94.2 \pm 3.3$ & $94.7 \pm 2.8$ & $82.6 \pm 6.1$ \\
    \hline
    H\&E + CD3, CD20, BCL1 & $83.7 \pm 6.2$ & $83.7 \pm 7.0$ & $94.4 \pm 3.3$ & $95.5 \pm 2.6$ & $83.4 \pm 6.1$ \\
    \hline
    H\&E + CD10, CD20, CD3, EBV ISH, BCL1, CD30 & $85.3 \pm 5.4$ & $84.5 \pm 7.0$ & $93.5 \pm 3.7$ & $95.7 \pm 2.7$ & $84.7 \pm 6.5$ \\
    \hline
    
\multicolumn{4}{c}{\vspace{0.5cm}}\\

    \hline
    \multirow{2}{*}{\textbf{Method}} & \multicolumn{5}{c}{\textbf{Per-Class F1}} \\ \cline{2-6}
    & \textbf{B-Cell} & \textbf{HL}  & \textbf{FL, MZL} & \textbf{MCL} & \textbf{T-cell} \\
    \hline
    Baseline: Best H\&E Model & $78.0 \pm 7.2$ & $72.3 \pm 12.8$ & $62.9 \pm 17.1$ & $75.9 \pm 15.5$ & $17.4 \pm 21.1$ \\
    \hline
    Baseline: CD10, CD20, CD3, EBV ISH, BCL1, CD30 & $78.7 \pm 7.2$ & $81.1 \pm 12.2$ & $9.1 \pm 9.1$ & $87.5 \pm 10.2$ & $87.2 \pm 9.8$ \\
    \hline
    Baseline: best model with all immunostains & $86.7 \pm 6.0$ & $92.3 \pm 7.7$ & $58.1 \pm 19.3$ & $93.3 \pm 6.7$ & $97.3 \pm 2.7$ \\
    \hline
    H\&E + CD20 & $85.0 \pm 6.3$ & $74.4 \pm 13.4$ & $66.7 \pm 15.7$ & $80.0 \pm 13.8$ & $74.3 \pm 14.6$ \\
    \hline
    H\&E + CD3 & $84.7 \pm 6.5$ & $76.2 \pm 13.3$ & $58.1 \pm 18.4$ & $75.9 \pm 15.0$ & $86.5 \pm 10.3$ \\
    \hline
    H\&E + CD3, CD20 & $87.7 \pm 4.5$ & $82.1 \pm 11.5$ & $66.7 \pm 15.4$ & $80.0 \pm 14.1$ & $89.5 \pm 8.1$ \\
    \hline
    H\&E + CD3, CD20, BCL1 & $85.7 \pm 6.2$ & $85.0 \pm 10.5$ & $64.9 \pm 15.7$ & $90.3 \pm 9.7$ & $91.9 \pm 8.1$ \\
    \hline
    H\&E + CD10, CD20, CD3, EBV ISH, BCL1, CD30 & $87.2 \pm 5.9$ & $84.2 \pm 10.2$ & $70.6 \pm 16.1$ & $90.3 \pm 9.7$ & $91.9 \pm 8.1$ \\
    \hline
\end{tabular}

\caption{Performance comparison of the best H\&E-only model, models using only immunostains, and models using features from H\&E combined with selected immunostains. All experiments are performed on the five-way grouped classification task. }  
\label{table8}
\end{table*}

\begin{table*}
\begin{tabular}{cc|c|cc|cccc|c}
\hline
\multicolumn{2}{c|}{\textbf{Comparison}} &
  \multirow{2}{*}{\textbf{\begin{tabular}[c]{@{}c@{}}Accuracy\\ Difference \\ (Method 1 -\\ Method 2)\end{tabular}}} &
  \multicolumn{2}{c|}{\textbf{\begin{tabular}[c]{@{}c@{}}Two-tailed\\ Paired t-test\\      (Bootstrapping)\end{tabular}}} &
  \multicolumn{4}{c|}{\textbf{\begin{tabular}[c]{@{}c@{}}Test for\\ Equivalence \\ (TOST with\\ Bootstrapping)\end{tabular}}} &
  \multirow{2}{*}{\textbf{\begin{tabular}[c]{@{}c@{}}Conclusion\\ ($\alpha=5\%$)\end{tabular}}} \\ \cline{1-2} \cline{4-9}
\multicolumn{1}{c|}{Method 1} &
  Method 2 &
   &
  \multicolumn{1}{c|}{\begin{tabular}[c]{@{}c@{}}95\%\\ CI\\ Lower\\ Bound\end{tabular}} &
  \begin{tabular}[c]{@{}c@{}}95\%\\ CI\\ Upper\\ Bound\end{tabular} &
  \multicolumn{1}{c|}{\begin{tabular}[c]{@{}c@{}}90\%\\ CI\\ Lower\\ Bound\end{tabular}} &
  \multicolumn{1}{c|}{$-\Delta$} &
  \multicolumn{1}{c|}{\begin{tabular}[c]{@{}c@{}}90\%\\ CI\\ Upper\\ Bound\end{tabular}} &
  $+\Delta$ &
   \\ \hline
\multicolumn{1}{c|}{\begin{tabular}[c]{@{}c@{}}Best\\ H\&E\\ Model\end{tabular}} &
  \begin{tabular}[c]{@{}c@{}}Hemato-\\ pathologist\\ 1 on TMAs\end{tabular} &
  0.082 &
  \multicolumn{1}{c|}{-0.030} &
  0.193 &
  \multicolumn{1}{c|}{-0.015} &
  \multicolumn{1}{c|}{-0.050} &
  \multicolumn{1}{c|}{0.173} &
  0.050 &
  Non-inferior \\ \hline
\multicolumn{1}{c|}{\begin{tabular}[c]{@{}c@{}}Best\\ H\&E\\ Model\end{tabular}} &
  \begin{tabular}[c]{@{}c@{}}Hemato-\\ pathologist\\ 2 on TMAs\end{tabular} &
  0.040 &
  \multicolumn{1}{c|}{-0.072} &
  0.155 &
  \multicolumn{1}{c|}{-0.056} &
  \multicolumn{1}{c|}{-0.050} &
  \multicolumn{1}{c|}{0.138} &
  0.050 &
  -- \\ \hline
\multicolumn{1}{c|}{\begin{tabular}[c]{@{}c@{}}Best\\ H\&E\\ Model\end{tabular}} &
  \begin{tabular}[c]{@{}c@{}}Hemato-\\ pathologist\\ 3 on WSIs\end{tabular} &
  0.005 &
  \multicolumn{1}{c|}{-0.119} &
  0.119 &
  \multicolumn{1}{c|}{-0.090} &
  \multicolumn{1}{c|}{-0.050} &
  \multicolumn{1}{c|}{0.102} &
  0.050 &
  -- \\ \hline
\multicolumn{1}{c|}{\begin{tabular}[c]{@{}c@{}}Best\\ H\&E\\ Model\end{tabular}} &
  \begin{tabular}[c]{@{}c@{}}General\\ Pathologist\\ on WSIs\end{tabular} &
  0.080 &
  \multicolumn{1}{c|}{-0.036} &
  0.197 &
  \multicolumn{1}{c|}{-0.021} &
  \multicolumn{1}{c|}{-0.050} &
  \multicolumn{1}{c|}{0.175} &
  0.050 &
  Non-inferior \\ \hline

\multicolumn{10}{c}{\vspace{0.5cm}}\\
\hline
\multicolumn{2}{c|}{\textbf{Comparison}} &
  \multirow{2}{*}{\textbf{\begin{tabular}[c]{@{}c@{}}Accuracy\\ Difference \\      (Method 1 -\\ Method 2)\end{tabular}}} &
  \multicolumn{2}{c|}{\textbf{\begin{tabular}[c]{@{}c@{}}Two-tailed\\ Paired t-test\\      (Bootstrapping)\end{tabular}}} &
  \multicolumn{4}{c|}{\textbf{\begin{tabular}[c]{@{}c@{}}Test for\\ Equivalence\\ (TOST with\\ Bootstrapping)\end{tabular}}} &
  \multirow{2}{*}{\textbf{\begin{tabular}[c]{@{}c@{}}Conclusion\\ ($\alpha=5\%$)\end{tabular}}} \\ \cline{1-2} \cline{4-9}
\multicolumn{1}{c|}{Method 1} &
  Method 2 &
   &
  \multicolumn{1}{c|}{\begin{tabular}[c]{@{}c@{}}95\%\\ CI\\ Lower\\ Bound\end{tabular}} &
  \begin{tabular}[c]{@{}c@{}}95\%\\ CI\\ Upper\\ Bound\end{tabular} &
  \multicolumn{1}{c|}{\begin{tabular}[c]{@{}c@{}}90\% \\ CI\\ Lower\\ Bound\end{tabular}} &
  \multicolumn{1}{c|}{$-\Delta$} &
  \multicolumn{1}{c|}{\begin{tabular}[c]{@{}c@{}}90\% \\ CI\\ Upper\\ Bound\end{tabular}} &
  $+\Delta$ &
   \\ \hline
\multicolumn{1}{c|}{\begin{tabular}[c]{@{}c@{}}Best\\ H\&E\\ Model\end{tabular}} &
  \begin{tabular}[c]{@{}c@{}}TripletNet\\ finetuned\\ (Camelyon)\end{tabular} &
  0.121 &
  \multicolumn{1}{c|}{0.003} &
  0.246 &
  \multicolumn{1}{c|}{--} &
  \multicolumn{1}{c|}{-0.050} &
  \multicolumn{1}{c|}{--} &
  0.050 &
  \begin{tabular}[c]{@{}c@{}}Significant\\ Difference\end{tabular} \\ \hline
\end{tabular}

\caption{Summary of statistical tests comparing the best H\&E-only model to pathologists and deep-learning models.}  
\label{table9}
\end{table*}

\begin{table*}
\begin{tabular}{cc|c|cc|cccc|c}
\hline
\multicolumn{2}{c|}{\textbf{Comparison}} &
  \multirow{2}{*}{\textbf{\begin{tabular}[c]{@{}c@{}}Accuracy \\ Difference \\ (Method 1 -\\ Method 2)\end{tabular}}} &
  \multicolumn{2}{c|}{\textbf{\begin{tabular}[c]{@{}c@{}}Two-tailed\\ Paired t-test\\ (Bootstrapping)\end{tabular}}} &
  \multicolumn{4}{c|}{\textbf{\begin{tabular}[c]{@{}c@{}}Test for\\ Equivalence\\ (TOST with\\ Bootstrapping)\end{tabular}}} &
  \multirow{2}{*}{\textbf{\begin{tabular}[c]{@{}c@{}}Conclusion\\ ($\alpha=5\%$)\end{tabular}}} \\ \cline{1-2} \cline{4-9}
\multicolumn{1}{c|}{\begin{tabular}[c]{@{}c@{}}Method\\ 1\end{tabular}} &
  \begin{tabular}[c]{@{}c@{}}Method\\ 2\end{tabular} &
   &
  \multicolumn{1}{c|}{\begin{tabular}[c]{@{}c@{}}95\% \\ CI\\ Lower\\ Bound\end{tabular}} &
  \begin{tabular}[c]{@{}c@{}}95\%\\ CI\\ Upper\\ Bound\end{tabular} &
  \multicolumn{1}{c|}{\begin{tabular}[c]{@{}c@{}}90\%\\ CI\\ Lower\\ Bound\end{tabular}} &
  \multicolumn{1}{c|}{$-\Delta$} &
  \multicolumn{1}{c|}{\begin{tabular}[c]{@{}c@{}}90\%\\ CI\\ Upper\\ Bound\end{tabular}} &
  $+\Delta$ &
   \\ \hline
\multicolumn{1}{c|}{\begin{tabular}[c]{@{}c@{}}6 Stains\\ Only\\ (CD10,\\ CD20,\\ CD3,\\ EBV ISH,\\ BCL1,\\ CD30)\end{tabular}} &
  \multirow{3}{*}{\begin{tabular}[c]{@{}c@{}}All 46\\ Stains\end{tabular}} &
  -0.112 &
  \multicolumn{1}{c|}{-0.209} &
  -0.016 &
  \multicolumn{1}{c|}{--} &
  \multicolumn{1}{c|}{-0.050} &
  \multicolumn{1}{c|}{--} &
  0.050 &
  \begin{tabular}[c]{@{}c@{}}Significant\\ Difference\end{tabular} \\ \cline{1-1} \cline{3-10} 
\multicolumn{1}{c|}{\begin{tabular}[c]{@{}c@{}}H\&E\\ Features\\ Only\end{tabular}} &
   &
  -0.173 &
  \multicolumn{1}{c|}{-0.271} &
  -0.078 &
  \multicolumn{1}{c|}{--} &
  \multicolumn{1}{c|}{-0.050} &
  \multicolumn{1}{c|}{--} &
  0.050 &
  \begin{tabular}[c]{@{}c@{}}Significant\\ Difference\end{tabular} \\ \cline{1-1} \cline{3-10} 
\multicolumn{1}{c|}{\begin{tabular}[c]{@{}c@{}}H\&E\\ Features\\ + 6 Stains\end{tabular}} &
   &
  -0.009 &
  \multicolumn{1}{c|}{-0.101} &
  0.070 &
  \multicolumn{1}{c|}{-0.085} &
  \multicolumn{1}{c|}{-0.050} &
  \multicolumn{1}{c|}{0.062} &
  0.050 &
  -- \\ \hline
\end{tabular}
\caption{Summary of statistical tests comparing different models (model with selected immunostains, model using features extracted from H\&E and selected immunostains) to a baseline model using all 46 available immunostains. }  
\label{table10}
\end{table*}

\begin{table*}[!htbp]
\centering
\begin{tabular}{c | c | c }
    \hline
    \textbf{Number of Patches per Core} & \textbf{Patch-Level CV Accuracy} & \textbf{Core-Level CV Accuracy} \\
    \hline
    1 & 54.1\% & 55.4\% \\
    4 & 52.4\% & 58.3\% \\
    9 & 48.7\% & 57.4\% \\
    16 & 46.8\% & 57.9\% \\
    25 & 44.9\% & 57.5\% \\
    36 & 43.3\% & 55.2\% \\
    49 & 42.7\% & 53.6\% \\
    64 & 41.6\% & 53.9\% \\
    81 & 40.9\% & 53.6\% \\
    100 & 40.8\% & 53.6\% \\
    \hline
\end{tabular}
\caption{There is no standard patch size \citep{Steinbuss} so we performed patch-resolution experiments to select the best patch size for feature-based models (at the extreme, using one patch to represent the core). We present patch-level and core-level cross-validation (CV) accuracies for the nuclei-only model trained using different numbers of patches per-core. We considered cases when each core was divided into a perfect square number of patches (1, 4, 9, …, 100 patches). Using this patch extraction method, we divided the width and height of each TMA core into a fixed number of segments to produce a grid of equally-sized patches. Since TMA cores come in varying sizes, this method preserves the same label distribution in the patch-level dataset as in the original core-level dataset as we simply scale up the number of examples by a constant. We compared models fitted using features aggregated from patches of different sizes, and selected the best model based on the 5-fold cross-validation accuracy. We found experimentally that extracting a small number of patches per core (specifically, 4 patches per core) led to the best model performance, and in particular, better performance than core-level model training and prediction. }  
\label{table11}
\end{table*}

\end{document}